\documentclass[10pt,journal,letterpaper,compsoc,twoside]{IEEEtran}
\usepackage{graphicx}
\usepackage{setspace}
\usepackage{times}
\usepackage{epsfig}
\usepackage{algo}
\usepackage{amsmath}
\usepackage{amssymb}
\usepackage{multirow}
\usepackage{url}
\usepackage{wrapfig}

\input{alldefin}

\begin{document}

\title{Scale Adaptive Clustering of Multiple Structures}

\author{Xiang~Yang~ and~  Peter~Meer  
\IEEEcompsocitemizethanks{\IEEEcompsocthanksitem
X. Yang (Dept. of Mechanical and Aerospace  Engineering),~
P. Meer (Dept. of Electrical and Computer Engineering), 
Rutgers University, NJ 08854, USA. \protect
E-mail: xiang.yang@yahoo.com, meer@soe.rutgers.edu}
\thanks{}}

\markboth{
Yang and Meer: Scale Adaptive Clustering of Multiple Structures}
{Yang and Meer: Scale Adaptive Clustering of Multiple Structures}

\IEEEcompsoctitleabstractindextext{
\begin{abstract}
We propose the segmentation of noisy datasets into Multiple Inlier Structures with a new Robust Estimator (MISRE).
The scale of each individual structure is estimated adaptively 
from the input data and refined by mean shift, 
{\it without} tuning any parameter in the process,
or manually specifying thresholds for different estimation problems.
Once all the data points were classified into separate structures, 
these structures are sorted by their densities with the strongest inlier structures coming out first.
Several 2D and 3D synthetic and real examples are presented to illustrate 
the efficiency, robustness and the limitations of the MISRE algorithm.

\end{abstract}

}

\maketitle
\IEEEdisplaynotcompsoctitleabstractindextext
\IEEEpeerreviewmaketitle

\section{Introduction}
\label{sec:introduction}

\IEEEPARstart{T}oday in computer vision, 
people use convolutional neural networks to identify 
complex patterns in natural settings. These methods give only
qualitative categorizations instead of parameterizations of the models.
The training process could take a significantly long time and the much smaller testing set has to be similar enough to those used for training.

When the input contains structures governed by precise mathematical relations, the objective functions can be estimated quantitatively 
without any pre-training. 
The estimator has to be robust to detect the inlier structures
while removing the structureless outliers.
The objective functions are either linear, 
like the estimation of 3D planes, or nonlinear, 
such as finding 3D spheres or the homography between two 2D images. 
The input dataset could also contain multiple inlier structures with every
structure corrupted by different noise.

The elemental subsets are the building blocks of the robust regression.
Each randomly chosen subset has the minimum number of input points
required to estimate the parameters in the objective function.
The most used algorithm for robust fitting in the past 35 years is
the RANdom SAmple Consensus (RANSAC) \cite{fischler81}. 
Similar types of methods also exist, 
like PROSAC, MLESAC, Lo-RANSAC, etc.
These algorithms use different ways to generate the random sampling
and/or probabilistic relations for the elimination of the outliers. 
In paper \cite{raguram08}, a review of these methods was given.

Before the estimation,
the user has to assign a value for the inlier noise level, 
which is a major drawback of using RANSAC.
Providing too small of a scale could filter out many inlier points, 
while too large of a value bring in outliers.
This threshold is not even mentioned sometime since
in real images the noise is often less than 3 pixels.
However, an appropriate value of the scale is hardly
predictable without prior knowledge.

The problem also appears in methods which begin with RANSAC,
like J-linkage \cite{Toldo08}.
When the images are resized, the error changes proportionally and 
the given scale also has to be changed to get the correct result. 
Methods using adaptive scale rather than hard thresholding, like
T-linkage \cite{Magri14}, 
may not work properly when different noise exists for
different inlier structures.

The inlier scale can also be predicted based on statistical distributions.
Based on \cite[Section 3.2.2]{Wand95}, 
the paper \cite{lavva08} combined the
projection based M-estimator \cite{chen03} with RANSAC 
to detect 3D geometric primitives,
but the input acquired by 3D laser scanner had low noise and 
did not contain randomly distributed outliers.
The assumption based on a statistical distribution is only valid 
for specific problems.

In \cite{raguram13}, a universal framework for RANSAC 
(USAC) was proposed,
where the threshold was based on Gaussian distribution of the inliers.
The method in \cite{litman15} performed better than USAC,
estimating the inlier rate from probabilistic reasoning,
instead of specifying the inlier scale. 
However, \cite{litman15} required a very dense sampling and only one inlier structure could be recovered.

Other methods tried to avoid the inlier threshold and
transformed the estimation into 
an energy-based minimization problem.
These approaches first identified a possible inlier region, 
and then iteratively improved the solution to get the final estimate.
In \cite{wang12} the $k$-th ordered absolute residual was 
sequentially improved.
The number of inlier structures had to be specified 
before the estimation \cite{elhamifar13}.
In \cite{tennokoon16} $p+2$ points were randomly selected to 
evaluate the error, where $p$ was the size of the elemental subset.
The significance of taking two additional points in a sample 
was not justified and the parameter $k$ varied largely in the experiments.

The Propose Expand and Re-estimate Labels (PEARL) algorithm in \cite{boykov12} started with RANSAC,
followed with alternative steps of 
expansion (inlier classification) and re-estimation to minimize the energy of the errors. A synthetic example showed that PEARL can handle the estimation of multiple 2D lines with different Gaussian noises,
but was not tested on any other models.
The amount of outliers was small in all the experiments.
The Random Cluster Model SAmpler (RCMSA) in \cite{pham14} was
similar to \cite{boykov12}, but simulated annealing was used to minimize the overall energy function. 

The generalized projection-based M-estimator (gpbM)  \cite{mittalanand12} tried to locate a dense region where an inlier structure could exist, by locating the highest value of the cumulative distribution function weighted by its size.
The assigned weights were critical to obtain the correct scale estimate, 
as Figure 3 from \cite{mittalanand12} showed. 
However, this weighting strategy cannot work all the time due to the interaction between all existing inliers and outliers.

\begin{figure}[t]
\includegraphics[scale=0.41]{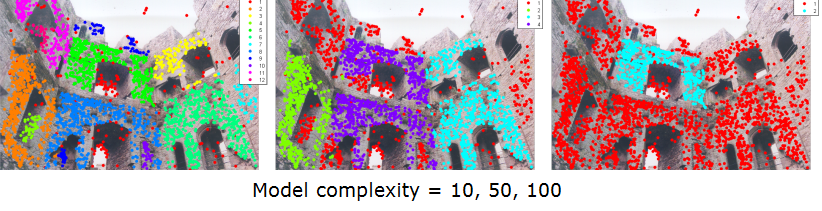}
\caption{Homography estimation in RCMSA \cite{pham14} 
with different model complexity values.}
\label{fig:rcmsa}
\vspace{-0.5cm}
\end{figure}

These methods also do not put emphasis on the internal parameter(s) 
which often have to be modified for a particular estimation task. 
For example, by varying the model complexity value in \cite{pham14},
the homography estimation showed similar behavior as 
changing the RANSAC inlier scale (Fig.\ref{fig:rcmsa}).
The default value was 100 for the fundamental matrix and 10 for homography. Applying the values vice-versa, the estimator no longer worked. This adjustment needs prior knowledge 
since there is no systematic way to predict the values of these parameters. 
To avoid {\it all} the internal parameters
and handle {\it different} structure scales, 
a general method for robust estimation should 
{\it independently} estimate the fitting error 
for each individual structure.

In this paper, we propose a new algorithm for 
the Multiple Inlier Structures Robust Estimator (MISRE).
The estimation for each structure consists of three consecutive steps: 
scale estimation, mean shift based structure recovery, and strength based inlier classification. 
The only parameter to be specified by the user is the number of trials for random sampling, which is required in any robust estimator using elemental subsets.
The major innovations of the paper are summarized below.
\begin{itemize}
\item Adaptive scale estimation for each individual structure.
\item No tuning of the internal parameters or threshold is required for different objective functions.
\item Structures are characterized by their strength (average density),
and in general an inlier structure has a stronger strength.
\item The efficiency, robustness and the limitations of the new robust 
estimator are discussed in detail.
\end{itemize}
Our experiments show successful results for different objective functions
without tuning any parameters, and the entire process is self-adaptive.

The linearized objective functions are presented in Section \ref{sec:prepare}. The algorithm of the new robust estimator is detailed in Section \ref{sec:newscale}.
Experiments of different estimation problems
are given in Section \ref{sec:experiments}.
Finally, in Section \ref{sec:discussion} 
we compare the new estimator with other methods and 
discuss some open problems.

\section{Structure Initialization} 
\label{sec:prepare}

In Section \ref{sec:transformation},
the nonlinear objective function of the inputs is 
transformed into a linear function of the carrier vectors.
The first order approximation of the covariance matrix
of these carriers is also computed.
Section \ref{sec:reducedistance} explains how 
the largest Mahalanobis distance of each input point
is taken into account when 
multiple carrier vectors are derived.

\vspace{-.4cm}
\subsection{Carrier Vectors}
\label{sec:transformation}

The nonlinear objective functions in computer vision can be transformed 
into linear relations in higher dimensions.
These linear relations, containing terms formed by the input measurements and their pairwise products, are called {\it carriers}.
Each relation gives a carrier vector.
For linear objective functions, such as plane fitting, 
the input variables and the carrier vector are identical.

For example, in the estimation of fundamental matrix,
the input data $\by$ are the point correspondences from two images 
$\left[ x\;\;y\;\;x^\prime\;\;y^\prime \right]^\top \in \mathbb{R}^4$, 
with $l=4$ dimensions. 
The objective function with noisy image coordinates is
\begin{equation}
\label{eqn:epipolar}
[x^\prime~~ y^\prime~~ 1] ~\bF~ [x~~ y~~ 1]^\top \simeq  0
\end{equation}
which gives a carrier vector $\bx \in \mathbb{R}^8$ 
containing $m=8$ carriers\\
$\bx=\left[x\;\;y\;\;x^\prime\;\;y^\prime\;\;x x^\prime\;\;
x y^\prime\;\; y x^\prime \;\;y y^\prime \right]^\top$ .
The linearized function of the carriers is
\begin{equation}
\label{eqn:linear}
\bx_{i}^\top \btheta - \alpha \simeq 0 ~~~ i = 1,\ldots,n_{in}
\end{equation}
where $n_{in}$ denotes the number of inliers.
Vector $\btheta \in \mathbb{R}^8$ 
and scalar intercept $\alpha$ 
derived from the $3\times 3$ matrix $\bF$ are to be estimated.
The constraint $\btheta^{\top} \btheta = 1$ eliminates the multiplicative 
ambiguity in (\ref{eqn:linear}). 

In the general case, several linear equations can be derived 
from a single input $\by_i$
\begin{equation}
\label{eqn:linearSolve}
\bx_{i}^{[c]\top} \btheta - \alpha \simeq 0 ~~~ 
c = 1,\ldots,\zeta ~~~ i = 1,\ldots,n_{in} 
\end{equation}
corresponding to $\zeta$ different carrier vectors $\bx^{[c]}$.
For example, the estimation of homography 
has $\zeta =2$ carrier vectors derived from
$x$ and $y$ image coordinates.

The Jacobian matrix is required for the first order approximation of the covariance of carrier vector.
From each carrier vector $\bx^{[c]}$, an $m\times l$ Jacobian matrix
$\bJ_{\scriptsize{\bx^{[c]}|\by}}$ is derived.
Each column of the Jacobian matrix contains the derivatives of the 
$m$ carriers in $\bx^{[c]}$ with respect to one measurement from $\by$.
The Jacobian matrices derived from linear objective
functions are not input dependent, 
while those derived from
nonlinear objective functions rely on the specific input point. 
The carrier vectors are {\it heteroscedastic} for nonlinear objective 
functions. For example, the transpose of the $8\times 4$ Jacobian matrix
of the fundamental matrix
\begin{equation}
\label{eqn:fundamcov}
\bJ_{\scriptsize{\bx_i|\by_i}}^\top = \left[
\begin{array}{@{\hspace{-0.03cm}}c@{\hspace{0.15cm}}c@
{\hspace{0.15cm}} c@{\hspace{0.15cm}}c@{\hspace{0.15cm}}c@
{\hspace{0.15cm}}c@{\hspace{0.15cm}}c@{\hspace{0.15cm}}c@
{\hspace{-0.03cm}}}
1 & 0 & 0 & 0 & x_i^\prime & y_i^\prime & 0 & 0 \\ 
0 & 1 & 0 & 0 & 0 & 0 & x_i^\prime & y_i^\prime \\ 
0 & 0 & 1 & 0 & x_{i} & 0 & y_{i} & 0 \\ 
0 & 0 & 0 & 1 & 0 & x_{i} & 0 & y_{i}  
\end{array} \right] 
\end{equation}
depends on $\by_i$. 

The $l\times l$ covariance matrix of the 
measurements $\sigma^2\bC_{\scriptsize{\by}}$ 
with $\det \bC_{\scriptsize{\by}} = 1$, 
has to be provided before estimation.
This is a chicken-egg problem, since the input points 
have not yet been classified into inliers and outliers.
A reasonable assumption is to set $\bC_{\scriptsize{\by}}$ as the identity 
matrix $\bI_{\scriptsize \by}$, if no additional information is given.
The input data are considered as independent and identically distributed, and contain
{\it homoscedastic} measurements with the same covariance.

The covariance of the carrier vector $\sigma^2 \bC_i^{[c]}$ is
computed from
\begin{equation}
\label{eqn:newcovariance}
\sigma^2\bC_i^{[c]} = \sigma^2 \bJ_{\scriptsize{\bx_i^{[c]}|
\by_i}} ~\bC_{\scriptsize{\by}} ~ \bJ_{\scriptsize{\bx_i^{[c]}|
\by_i}}^\top 
\end{equation}
where the dimensions of $\bC_i^{[c]}$ is $m\times m$.
The scale $\sigma$ of the structure is unknown and to be estimated.

\subsection{Computation of the Mahalanobis Distances}
\label{sec:reducedistance}

The elemental subset needs $m_e = \lceil \frac{m}{\zeta} \rceil$ 
input points to uniquely define $\btheta$ and $\alpha$ in the linear space.
For example, the homography $(m = 8,\; \zeta = 2)$ requires four point 
pairs, and eight pairs are necessary for 
the fundamental matrix $(m = 8,\; \zeta = 1)$ 
if the 8-point algorithm is used.
The input data should be normalized \cite[Section 4.4]{hartley04},
and the obtained structures mapped back onto the original space.

For each $\btheta$, 
every carrier vector is projected to a scalar value
$z_i^{[c]} = \bx_i^{[c]\top} \btheta ,\; c=1,\ldots,\zeta$.
The average projection of the $m$ vectors
from an elemental subset is $\alpha$.
The variance of $z_i^{[c]}$ is 
$\sigma^2 H_i^{[c]} = \sigma^2\btheta^\top \bC_i^{[c]} \btheta$.

The Mahalanobis distance, scaled by an unknown $\sigma$,  indicating how far is a projection $z_i^{[c]}$ 
from $\alpha$, is computed from
\begin{eqnarray}
\label{eqn:distances} \nonumber
d_i^{[c]} &=& \sqrt{\left( \bx_i^{[c]\top} \btheta  -
\alpha \right)^\top \left(H_i^{{[c]}}\right)^{-1} \left( 
\bx_i^{[c]\top} \btheta - \alpha \right)} \\
&=& \frac{|\bx_i^{[c]\top} \btheta  - \alpha |}
{\sqrt{\btheta^\top \bC_i^{[c]} \btheta}} \qquad c=1,\ldots,\zeta .
\end{eqnarray} 
Each input point $\by_i$ gives a 
$\zeta$-dimensional Mahalanobis distance vector
\begin{equation}
\label{eqn:mahavector}
\bd_i = \left[ \, d_i^{[1]} ~\ldots ~d_i^{[\zeta]} \, \right]^\top
\qquad i=1,\ldots, n .
\end{equation}
The worst-case scenario is taken to retain the largest Mahalanobis 
distance $d_i^{[\tilde{c}_i]}$ from all the $\zeta$ values
\begin{equation}
\label{eqn:jmax}
\tilde{c}_i = \operatorname* {arg\,max}_{c=1,\ldots,\zeta} \; d_i^{[c]}.
\end{equation}
For different $\btheta$-s, the same input point may 
have its largest distance computed from different carriers.

The symbols related to the largest Mahalanobis distance are:
$\tilde{d}_i$, the largest Mahalanobis distance for input $\by_i$;
$\tilde{\bx}_i$, the corresponding $m\times 1$ carrier vector;
$\tilde{z}_i$, the scalar projection of $\tilde{\bx}_i$;
$\widetilde{\bC}_i$, the $m\times m$ covariance matrix of $\tilde{\bx}_i$;
$\widetilde{H}_i$, the variance of $\tilde{z}_i$; 
$\hat\sigma$, the scale multiplying $\widetilde{\bC}_i$ 
and $\widetilde{H}_i$, which has to be estimated.

\section{Estimation of Multiple Structures} 
\label{sec:newscale}

The Multiple Inlier Structures Robust Estimator (MISRE) 
is introduced in this section.
The scale $\hat\sigma$ 
for a structure is estimated in Section \ref{sec:newsigma} by an
expansion criteria. The estimated scale is used in the mean shift 
to re-estimate the structure in Section \ref{sec:newstructure}. 
The iterative process continues until not enough 
input points remain
for a further estimation. In Section \ref{sec:strength}, 
all the estimated structures are ordered by strengths with  
the strongest inlier structures returned first.
The limitations of the method are explained in Section \ref{sec:inlieroutlier} and
the conditions for robustness is discussed in Section \ref{sec:robustness}.

\subsection{Scale Estimation of A Structure}
\label{sec:newsigma}

Assume that $n$ input points remain in the current iteration.
The estimation process initializes $M$ random elemental subsets, 
each giving a $\btheta$ and an $\alpha$.

For every point $\by_i$, compute the largest Mahalanobis distance 
$\tilde{d}_i$
\begin{equation}
\label{eqn:orderdistances}
\tilde{d}_i = \frac{|\tilde{\bx}_i^\top \btheta  -
\alpha |}{\sqrt{\btheta^\top \widetilde{\bC}_i \btheta}} \ge 0 \qquad 
i=1,\ldots,n.
\end{equation} 
Sort the $n$-distances in ascending order, denoted 
$\tilde{d}_{[i]}$. 
In total $j=1,\ldots, M$ sorted sequences $\tilde{d}_{[i]j}$ are found.

Let $n_{\epsilon} \ll n$ represent a small amount of points
\begin{equation}
\label{eqn:ntotal}
n_{\epsilon} = \frac{\epsilon \, n}{100} \qquad 0 < \epsilon \ll 100
\end{equation}
where $\epsilon$ defines the size of $n_{\epsilon}$ in percentage
of the input amount.

Among all $M$ trials, find the single sequence 
that gives the {\it minimum sum} of Mahalanobis distances
from its first $n_{\epsilon}$ points
\begin{equation}
\label{eqn:minmaha}
\min_{M} \sum_{i=1}^{n_{\epsilon}} \tilde{d}_{[i]j}.
\end{equation}
The sequence containing $n$ points, denoted $\tilde{d}_{[i]_M}$,
is retained and
the first $n_{\epsilon}$ points form the {\it initial set}.

If inlier structures still exist and $M$ is sufficiently large, 
the points in the initial set
have a high probability to be selected
from a dense region in a single inlier structure.
The corresponding structure defining $\tilde{d}_{[i]_M}$,
though not accurate enough for the final estimate, 
is closely aligned with an inlier structure.

Two rules should be considered for the ratio $\epsilon\%$. 
First, ${n}_{\epsilon}$ should be smaller than the size of any inlier
structure to be estimated. 
Therefore, a small ratio is preferred to detect all the potential structures.
All our experiments start with $\epsilon\% = 5\%$.
The second rule is to have the size of ${n}_{\epsilon}$ at least five
times the number of points in the elemental subset,
as suggested in \cite[page 182]{hartley04}.
This rule of thumb reduces unstable results when relatively
few input points are provided.

\begin{figure*}
\centering
\begin{tabular}{@{\hspace{-0.0cm}}c@{\hspace{-0.0cm}}c
@{\hspace{-0.0cm}}c@{\hspace{-0.0cm}}c@{\hspace{-0.0cm}}c}
\includegraphics[scale=0.215]{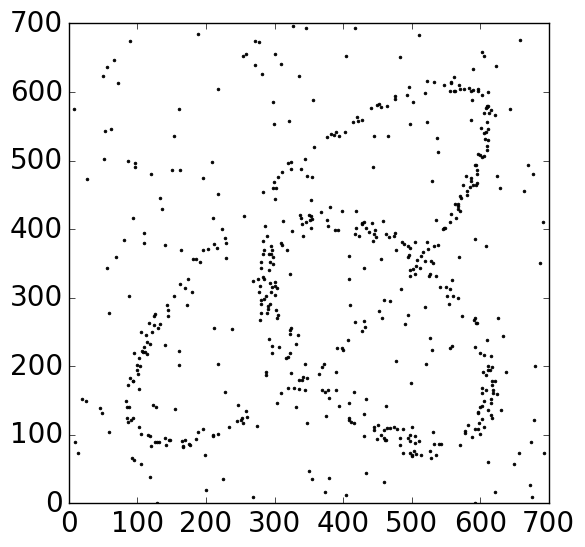}&
\includegraphics[scale=0.215]{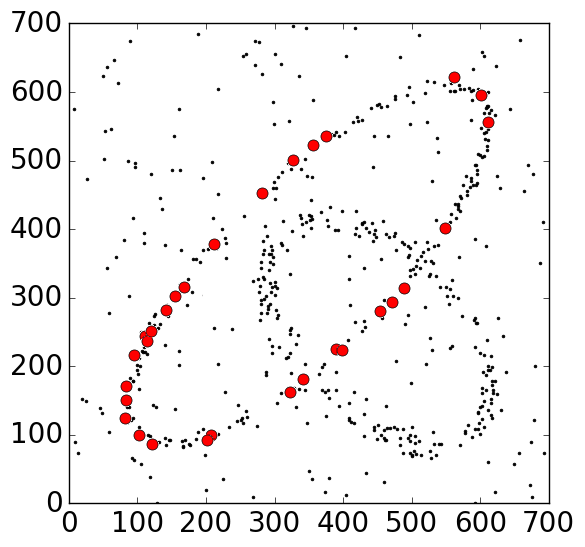}&
\includegraphics[scale=0.195]{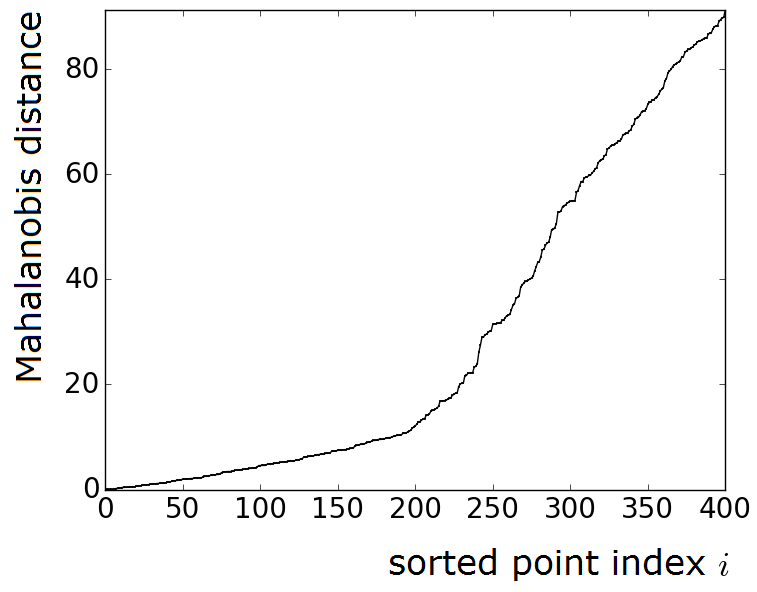}&
\includegraphics[scale=0.195]{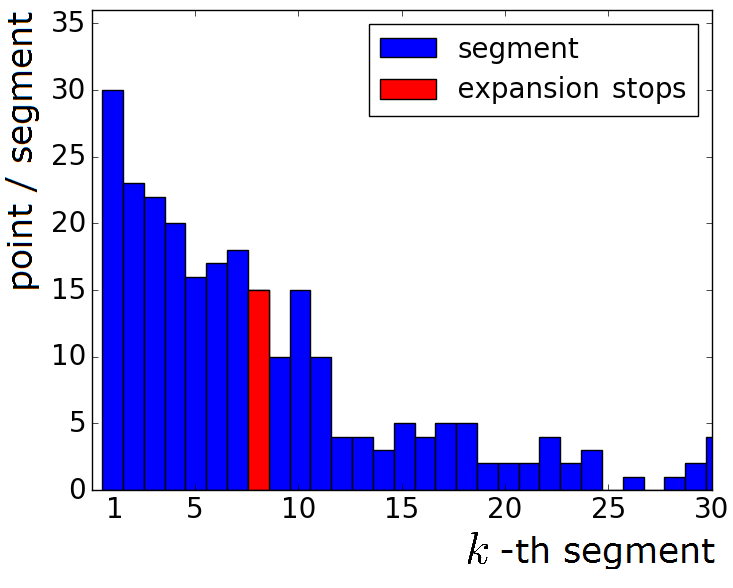}&
\includegraphics[scale=0.195]{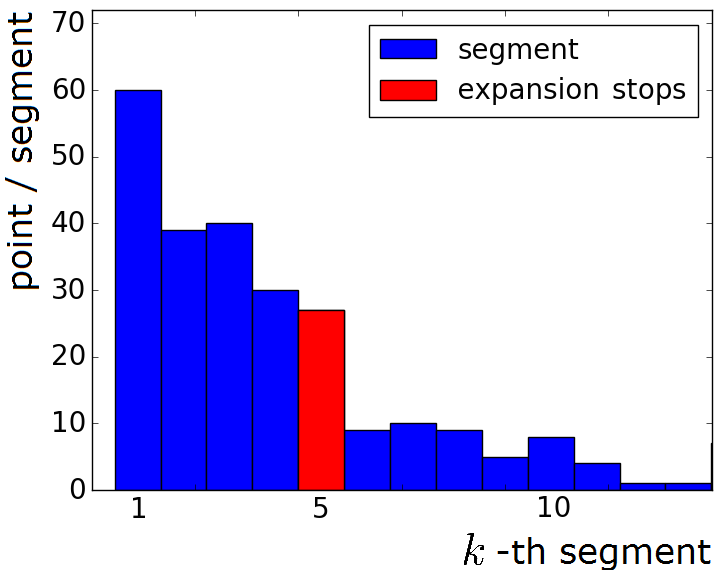}\\
(a) & (b) & (c) & (d) & (e)
\end{tabular}
\caption{Scale estimation.
(a) Input data.
(b) The initial set consists of ${n}_{\epsilon}$ points.
(c) The Mahalanobis distances of the first 400 points in 
$\tilde{d}_{[i]_M}$.
(d) Expansion with $\Delta d_5$.
(e) Expansion with $\Delta d_{10}$.}
\label{fig:computedistance}
\vspace{-.5cm}
\end{figure*}

\begin{figure}[t]
\begin{center}
\includegraphics[scale=0.3]{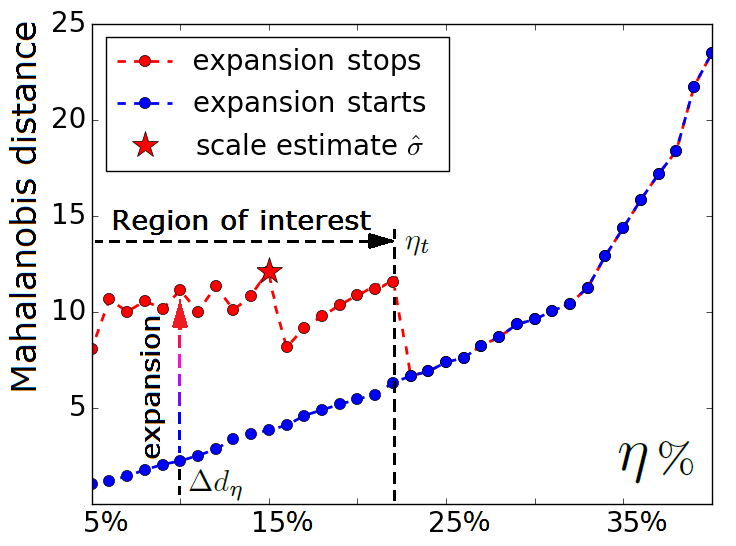}
\end{center}
\caption{Scale estimation using the Mahalanobis distances.
The expansion criteria applied to increasing sets.}
\label{fig:expand}
\vspace{-0.5cm}
\end{figure}

As many as possible points belonging to the same structure is 
needed to recover the scale $\hat \sigma$.  
The following example justifies the use of
expansion criteria for scale estimation.
In Fig.\ref{fig:computedistance}a two ellipses are shown; 
each of them has $n_{in} = 200$ inlier points.
They are corrupted by 
different Gaussian noise with standard deviation 
$\sigma_g = 5$ and $10$ respectively. Another
$n_{out} = 200$ outliers are randomly placed in the $700\times 700$ image.
With $M = 2000$ we obtain the sequence $\tilde{d}_{[i]_M}$, 
and its first ${n}_{\epsilon} = 30$ points are shown 
in Fig.\ref{fig:computedistance}b.
The sorted Mahalanobis distances in $\tilde{d}_{[i]_M}$ for the first 400 
points from 600 in total, are shown in Fig.\ref{fig:computedistance}c.

Divide the sequence $\tilde{d}_{[i]_M}$ into multiple segments,
and each segment covers an equal range of Mahalanobis distance 
$\Delta d_{\eta}$.
This corresponds to the Mahalanobis distance of the point
at $\eta\%$ position in the sequence $\tilde{d}_{[i]_M}$.
Let $n_{k}$ denote the number of points within 
the $k$-th segment, $k = 1, 2, 3 \ldots$, with $n_1= n_{\epsilon}$.
The expansion process verifies the following condition 
for each $k$
\begin{equation}
\label{eqn:condition} 
\frac{n_{k+1}}{\frac{1}{k}\sum_{i = 1}^{k}{n_i}} \leq 0.5
\end{equation}
where the numerator is the number of points
in the ($k+1$)-th segment and the denominator is the average point numbers inside all the $k$ segments. 

When the point density drops below half of 
the average in the previous segments, 
the boundary to separate this structure
from the outliers is found, $k=k_t$.
The value of the scale estimate is $k_{t}\Delta d_{\eta}$.
The condition (\ref{eqn:condition}) is a heuristic criteria
since the true distribution of the inliers is unknown.

Due to the randomness of the input data, 
the single scale estimate may not be stable 
and several independent expansions have to be generated
with different values of $\Delta d_{\eta}$.
This is a similar process 
to the Gaussian smoothing used in the Canny edge detection,
and the SIFT \cite{lowe04} over the scale space,
where discretization effect occurs when kernel size changes.
In Fig.\ref{fig:computedistance}d the expansion 
starting with $\Delta d_\epsilon = \Delta d_5$, 
stops at $k_{t_5}=8$, 
giving $\hat\sigma = 8.06$ (red bar).
In Fig.\ref{fig:computedistance}e the expansion with a larger $\Delta d_{10}$ stops at $k_{t_{10}}=5$ giving $\hat\sigma = 11.10$.

To stabilize the scale estimation, 
the expansion process is applied to an increasing sequence of sets.
As illustrated in Fig.\ref{fig:expand}, 
$\Delta d_{\eta}$ starts from 
$\Delta d_\epsilon = \Delta d_5$,
and increases by $1\%$ each time, 
$\eta = \epsilon, \epsilon + 1, \epsilon +2, \ldots$.
The blue points indicate the length of $\Delta d_{\eta}$.
Every expansion process is performed independently,
and stops at the corresponding red point 
when condition (\ref{eqn:condition}) is met.
The length of $\Delta d_{\eta}$ continues to increase until
it reaches a bound $\eta_t$,
where the following expansion with $\Delta d_{\eta_t + 1}$ can no longer expand beyond $k_t = 1$, as it is $22\%$ in Fig.\ref{fig:expand}. 

The scale estimate is found from the {\it region of interest}
where the sets of points can expand.
In Fig.\ref{fig:expand} it ranges from $5\%$ to $22\%$.
The region of interest may not always start from 
$\Delta d_{\epsilon}$, 
but at the first place where the expansion process begins, 
as $\Delta d_{\eta}$ increases.

The largest estimate from the region of interest 
gives the scale $\hat\sigma$
\begin{equation}
\label{eqn:estimatescale}
\hat\sigma = \max_{\eta=\epsilon,\ldots,\eta_t}  k_{t_\eta} 
\Delta d_\eta 
\end{equation}
the farthest expansion inside the region of interest. 
In Fig.\ref{fig:expand} the scale estimate is $\hat\sigma = 12.54$,
which is around $2\sim3\sigma_g$.

\vspace{-.2cm}
\subsection{Mean Shift Based Structure Recovery}
\label{sec:newstructure}

From the sequence $\tilde{d}_{[i]_M}$, collect
all the points within the scale estimate $\hat\sigma$.
Another $N \ll M$ elemental subsets are generated,
where the points are selected {\it only} from the collected subset.
We set $N = M/10$ since most points in this set come from the same structure.

For each trial {\it all} the remaining input points 
are projected by $\btheta$ to a one-dimensional space
$\tilde{z}_i = \tilde{\bx}_i^\top \btheta ,\; i=1,\ldots,n$.
The mean shift  \cite{comaniciu02} moves the
$z$ from $z = \alpha$ to the {\it closest mode} 
\begin{equation}
\label{eqn:kde}
\left[\widehat{\btheta},\widehat{\alpha}\right] 
= \operatorname*{arg\,max}_{\mathbf{\btheta},\alpha}
\frac{1}{n \hat\sigma} \sum_{i=1}^{n} \kappa \left(\left( z - 
\widetilde{z}_i \right)^\top\widetilde{B}_i^{-1} \left( z - \widetilde{z}_i
\right) \right).
\end{equation}
The variance $\tilde{B}_i$ is computed from
\begin{equation}
\label{eqn:oldfull}
\widetilde{B}_i = \hat{\sigma}^2 \widetilde{H}_i = 
\hat{\sigma}^2 {\btheta}^\top {\tilde\bC}_i {\btheta} 
= \hat{\sigma}^2 {\btheta}^\top \bJ_{\scriptsize{\tilde{\bx}_i|
\by_i}} \bJ_{\scriptsize{\tilde{\bx}_i|\by_i}}^\top {\btheta} 
\end{equation}
with $\bC_{\by} = \bI_{\by}$.

The function $\kappa(u)$ is the profile of a 
radial-symmetric kernel $K(u^2)$ defined only for $u \geq 0$. 
For the Epanechnikov kernel 
\begin{equation}
\label{eqn:profile}
\kappa(u) = \left\{
\begin{array}{rcr}
1 - u &  & \left( z - \widetilde{z}_i \right)^\top\widetilde{B}_i^{-1} 
\left( z - \widetilde{z}_i \right) \leq 1 \  \\
0 &  & \left( z - \widetilde{z}_i \right)^\top\widetilde{B}_i^{-1} 
\left( z - \widetilde{z}_i \right) > 1 .\\
\end{array} \right.
\end{equation}
Let $g(u) = -\kappa'(u)$ and for the Epanechnikov kernel, $g(u) = 1$ when
$0 \le u \leq 1$ and $0$ if $u > 1$.
All the points inside the window contribute equally in the mean shift. 
The convergence to the closest mode is obtained by assigning zero to the 
gradient of (\ref{eqn:kde}) in each iteration.
The $z_{new}$ is updated from the current value $z = z_{old}$ by
\begin{eqnarray}
\label{eqn:origmean}
z_{new}= \left[\sum_{i=1}^{n} {g\left(u\right)} \right]^{-1} \!\!
\left[\sum_{i=1}^{n}
{g\left(u\right)} \widetilde{z}_i \right] .
\end{eqnarray}
Many of the $n$ input points have their projections 
more distant from $z_{old}$ than $\pm\widetilde{B}_i$ and their 
weights are zeros.

\begin{figure}
\centering
\begin{tabular}{@{\hspace{-0.0cm}}c@{\hspace{-0.0cm}}c}
\includegraphics[scale=0.26]{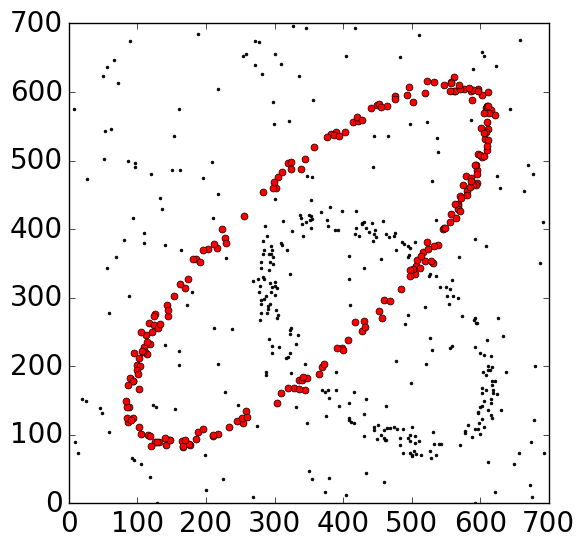}&
\includegraphics[scale=0.26]{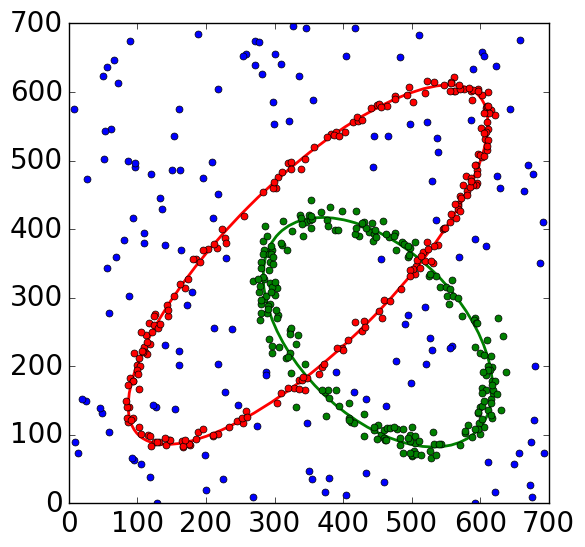}\\
(a) & (b)
\end{tabular}
\caption{Structure recovered and ordered by strengths.
(a) The first structure recovered after mean shift.
(b) Three recovered structures ordered by their strengths.}
\label{fig:strength}
\vspace{-0.5cm}
\end{figure}

The highest mode among all $N$ trials gives the estimate 
$\hat z = \hat\alpha$. The vector $\hat\btheta$ is obtained
from the same elemental subset which gives the highest mode 
$\hat\alpha$.
All the input points that can converge into the
$\pm \hat\sigma$ region around $\hat\alpha$ are 
classified as inliers, resulting in $n_{in}$ points. 
The total least squares (TLS) estimate for the structure is then 
computed to obtain $\hat\btheta^{tls}$, 
$\hat\alpha^{tls}$ and ${\hat\sigma}^{tls}$.
Fig.\ref{fig:strength}a shows the structure recovered in the 
first iteration, where most inlier points are collected compared with 
the result in Fig.\ref{fig:computedistance}b.

\vspace{-.3cm}
\subsection{Strength Based Classification}
\label{sec:strength}

After the mean shift step, the $n_{in}$ points 
are removed from the inputs before the next iteration. 
If the amount of the remaining data is smaller than $n_{\epsilon}$,
the algorithm terminates and all the recovered structures 
are sorted by their strengths in descending order.
The {\it strength of a structure} is defined as
\begin{equation}
\label{eqn:strength}
s = \frac{n_{in}}{\hat \sigma^{tls}}.
\end{equation}
which can also be seen as the density in the linear space of that structure.

Structures with stronger strengths are detected first,
and in general are inlier structures with more dense points
and smaller scales.
The new method does not rely on a threshold to separate 
inliers from the outliers, or assume any upper/lower limit to bound 
the range of error.
In Fig.\ref{fig:strength}b, three structures are returned
\[\begin{array}{rccc}
 & red  & green & blue\\
scale: & 16.4  & 35.4 & 708.7\\
inliers: & 219 & 210 & 163\\
strength: & 13.32 & 5.93 & 0.23
\end{array} \]
where the first two (red and green) are inlier structures.

If the scale estimator locates a structure consisting of outliers,
$\hat\sigma$ in general is much larger and
the strength weaker than inlier structures.
In real images, the difference in scale and strength between the 
inliers and the outlier is obvious to notice,
and the user can easily retain the inlier structures,
as the examples in Section \ref{sec:experiments} will show.
When an ambiguous inlier/outlier threshold appears, like in Fig.\ref{fig:synthellipse}c,
the strongest inlier structures are still detected correctly.

\vspace{-.3cm}
\subsection{Limitations}
\label{sec:inlieroutlier}

The major limitation of every robust estimator
comes from the interactions between inliers and outliers.
As the outlier amount increases, 
and/or the inlier structures become noisier,
eventually the inliers and outliers become less separable 
in the input space.
In our algorithm most of the processing is done in a linear space,
but the limitation introduced by outliers still exists.
We will illustrate it in the following example.

\begin{figure}[t]
\centering
\begin{tabular}{@{\hspace{-0.0cm}}c@{\hspace{-0.0cm}}c}
\includegraphics[scale=0.26]{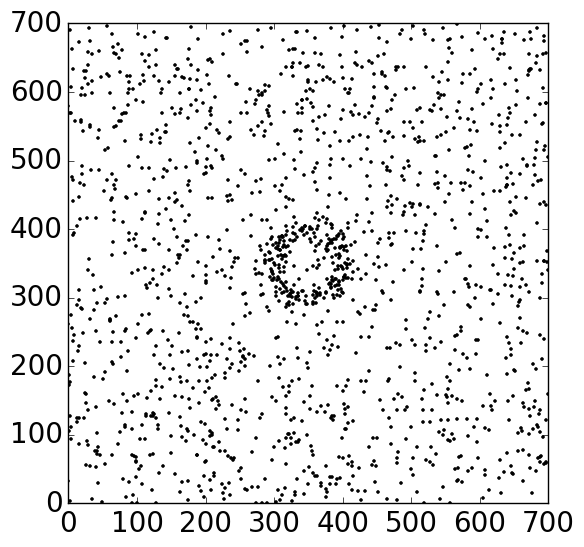}&
\includegraphics[scale=0.26]{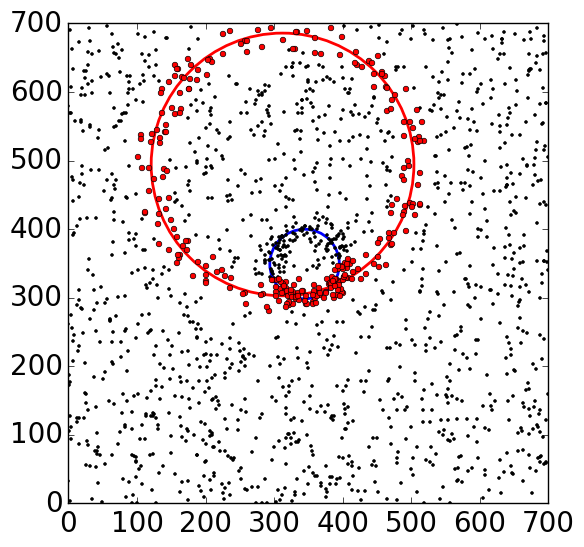} \\
(a) & (b) \\
\includegraphics[scale=0.26]{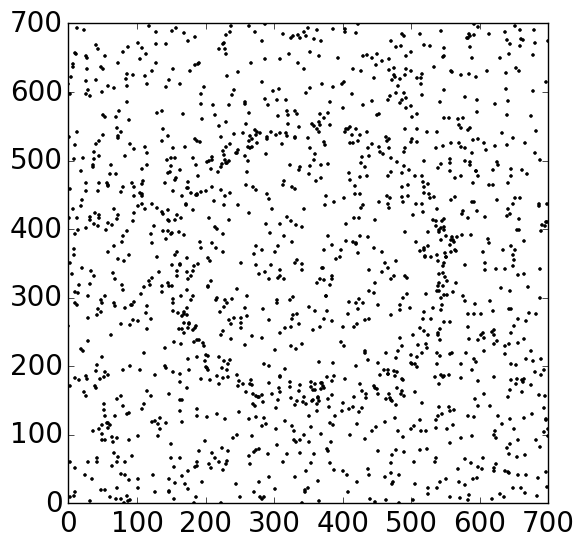}&
\includegraphics[scale=0.26]{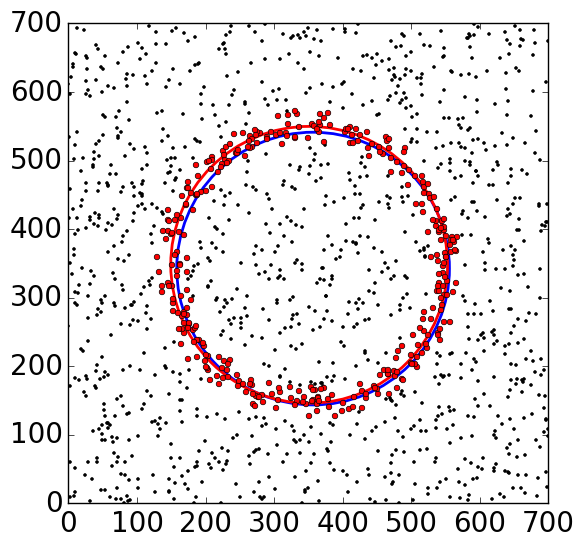} \\
(c) & (d)
\end{tabular}
\caption{The inlier/outlier interaction.
(a) Input data of a circle with radius 50. 
(b) Incorrect final result obtained 
from a correct scale estimate.
(c) Input data of a circle with radius 200. 
(d) Good final result obtained 
from a correct scale estimate.}
\label{fig:outlierproblem}
\vspace{-.5cm}
\end{figure}

In a $700\times 700$ image, a circle consists of $n_{in} = 200$ inliers is
corrupted by Gaussian noise with $\sigma_g = 10$, together with 
$n_{out} = 1500$ outliers.
The circles have different radiuses, 
50 in Fig.\ref{fig:outlierproblem}a and 
200 in Fig.\ref{fig:outlierproblem}c.
In both these figures,
the estimator finds the correct scale estimates
from the structure (blue circles) corresponding to the initial set,
where $\hat\sigma_{50} = 23.65$ and  $\hat\sigma_{200} = 23.58$.

About 196 true inlier points should exist inside the 
scale $\hat\sigma_{50} = 23.65$, 
based on the Gaussian distribution.
The number of outliers can be roughly estimated as 
\[ \quad (2\pi\, 50) (2*23.65) \frac{1500}{700\times 700}
= 45~ \mbox{points} \] 
giving 241 points in total inside the true inlier region.
However, after the mean shift step an incorrect final result (red circle)
containing 261 points is obtained in Fig.\ref{fig:outlierproblem}b,
where 84 points are true inliers and 177 points from the outliers.
Although the true structure appears more dense in the input space,
the mean shift converges
to an incorrect mode due to the heavy noise from outliers.

The circle in Fig.\ref{fig:outlierproblem}c appears
much weaker, however after 100 tests with randomly generated data 
(inlier/outlier), it returns more stable estimations than the smaller 
circle in Fig.\ref{fig:outlierproblem}a.
In a result shown in Fig.\ref{fig:outlierproblem}d,
346 points are classified as inliers, 
where 190 points are from true inliers
and 156 points from the outliers. The mean shift 
has a much lower probability to converge to another, incorrect mode,  
and this inlier structure resists more outliers.

Similar limitation exists in RANSAC when many outliers 
are present. Even a correct scale given by the user
can still lead to an incorrect estimation. 
The methods proposed in \cite{pham14} and \cite{tennokoon16}
returned incorrect results if too many outliers existed.
The failure of RANSAC also occurs due to not explicitly considering
the underlying task \cite{hassner14}.

In \cite{vedaldi05} a robust estimator for structure from motion algorithm was proposed combining an extended Kalman filter.
The example in Figure 5 returned correct estimate with the
data containing more than 60\% outliers.
For the homography estimation of Figure 6 in [22], 
more than 90\% of the points were outliers and correct result 
was still obtained.
However, none of these results had repetitive tests, and
the stability of the methods cannot be verified.

In MISRE, the strength of an inlier structure
is another factor with a strong influence on
the inlier/outlier interaction.
Firstly, the level of the inlier noise $\hat\sigma$ 
affects the number of outliers that can be tolerated.
With the same number of inliers,
structures with lesser inlier noise will have its
initial sets better aligned with the true structures,
and result in more reliable scale estimate.
Noisier inlier structures are most likely to interact with 
the outliers and lead to spurious results, see Fig.\ref{fig:synthellipse}d.

\begin{figure}[t]
\centering
\begin{tabular}{@{\hspace{-0.0cm}}c@{\hspace{-0.0cm}}c}
\includegraphics[scale=0.22]{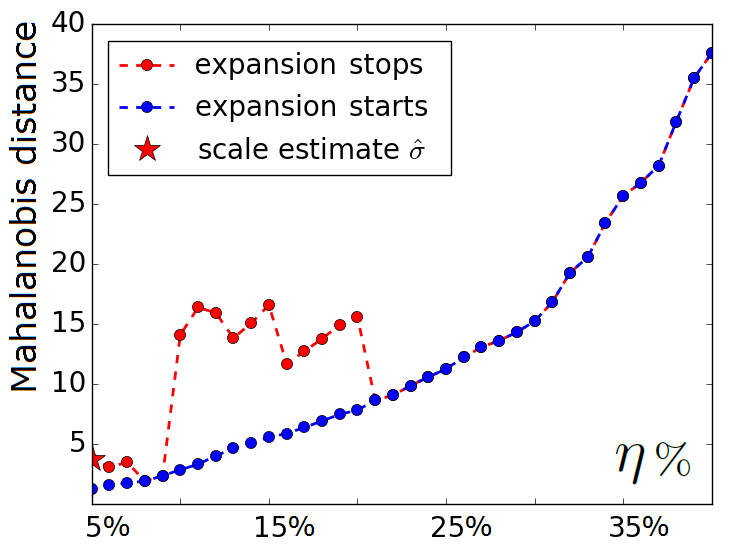}&
\includegraphics[scale=0.22]{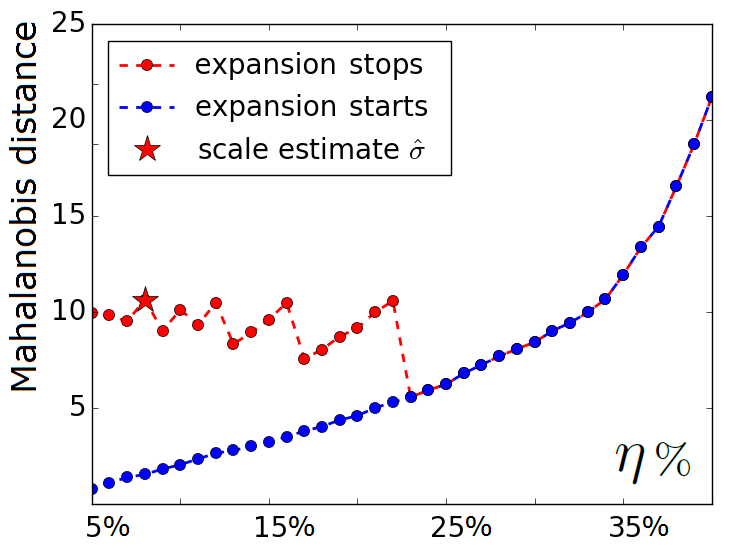} \\
(a) & (b)
\end{tabular}
\caption{Limitation of scale estimation.
(a) Unstable estimate obtained 
from a small number of inliers ($n_{in} = 200, 
n_{out} = 400$). 
(b) More robust estimate obtained 
from a larger number of inliers ($n_{in} = 400, 
n_{out} = 400$).}
\label{fig:limitproblem}
\vspace{-.5cm}
\end{figure}

In Fig.\ref{fig:limitproblem}a,
the expansion process is applied to
the same example as in Fig.\ref{fig:computedistance}a, but
with $n_{out} = 400$.
The expansion stops soon and the algorithm
locates the region of interest between $5\%-7\%$
giving a small scale estimate. 
After applying the expansion criteria
many times,
the range of expansion from different $\Delta d_{\eta}$
are not stable.
A too small value of the scale may attract only a 
minority of the inlier points.
In Fig.\ref{fig:limitproblem}b the number of inliers 
is raised to $n_{in} = 400$.
The scale estimate becomes a more stable value with
the region of interest located between $5\% - 22\%$. 

Secondly, when the inlier point amount $n_{in}$ is relatively small
compared to the outliers,
the region of interest also becomes unstable and
the initial set may not closely align with a true structure.
The expansion process then will not return a correct scale estimate.

When the inlier/outlier interaction is strong,
preprocessing on the input data is required to 
obtain more inlier points, and/or reduce the outlier amount
for a better performance.
In Fig.\ref{fig:homographyFig1}a of Section \ref{sec:experiments}, 
an example is given where homography estimation in 2D
is used to segment objects in 3D scene.
Under the small translational motions, the two planes on the bus
though orthogonal in 3D, are not separable in 2D 
due to the relatively small amount of inlier points.
In Fig.\ref{fig:homographyFig1}d we show that by using more inlier points, the estimator will recover more inlier structures.

If an inlier structure appears split in several structures
with fewer points, post-processing is needed to merge them.
The user can easily locate them by their strengths
since most of these split structures are still
stronger compared with the outliers.
The similarity of two structures should be compared
in the input space where measurements are obtained,
as the derived carriers in the linear space
do not represent the nonlinearities of the inputs explicitly.

For two inlier structures with linear objective function, 
the merge can be implemented based on the 
orientation of each structure and the distance between them.
For two ellipses, the geometric tools to determine the overlap area 
can be used \cite{hughes12}.
The 2D measurements of fundamental matrices and the
homographies are in the projective space instead of euclidean,
and the reconstructed 3D information should be applied to separate or merge the similar structures.

\subsection{Conditions for robustness}
\label{sec:robustness}

The number of trials $M$ for random sampling is the 
only parameter given by the user.
The required amount of $M$ depends strongly on the data 
to be processed.
The complexity of the objective function,
the size of the input data, the number of inlier structures, 
the inlier noise levels, and the amount of outliers,
all are factors which can affect the required number of trials.

If no information on the size of $M$ is known, 
the user can run several tests with different $M$-s until the results become stable. 
When the interaction between inliers and outliers is apparent,
the quality of the estimation cannot be compensated by a larger $M$ since the initial set has become less reliable. 
Only through preprocessing of the input data will the number of inlier points increase and a better result be obtained.

Three main conditions to improve 
the robustness of the proposed algorithm are summarized:
\begin{itemize}
\item Preprocessing to reduce the amount of outliers, 
while bring in more inliers.
\item The sampling size $M$ should be large enough to stably find the inlier estimates.
\item Post-processing should be done in the input space
when an inlier structure comes out split or has to be separated.
\end{itemize} 

\subsection{Review of the algorithm}
\label{sec:flowchart}

The MISRE algorithm is summarized below.

\noindent\hrulefill\\
{\bf \centerline{
Scale adaptive clustering of multiple structures}}
{\vspace{0.01cm}
\noindent\hrulefill}\\
\noindent{\bf Input}:
$\by_i$, $i=1,\ldots,n$ data points that contain an unknown number of 
inlier structures with their scales unspecified, along with outliers.
The covariance matrices for $\by_i$ are 
$\bC_{\scriptsize\by}=\bI_{\scriptsize\by}$ 
if not provided explicitly.

\noindent{\bf Output}:
The sorted structures with inliers come out first.

\begin{itemize}
\item Compute the carriers $\bx_i^{[c]}$, $c=1,\ldots,\zeta$, and the 
Jacobians $\bJ_{\scriptsize{\bx_i^{[c]}|\by_i}}$, for each input
$\by_i$,  $i=1,\ldots,n$.

\vspace{0.3cm}
\item[$\odot$] Generate $M$ random trials 
based on elemental subsets. 
\begin{itemize}
\item For each elemental subset find $\btheta$ and $\alpha$.
\item Compute the Mahalanobis distances from $\alpha$ for all carrier
vectors $\bx_i^{[c]}$, $c=1,\ldots,\zeta$. Keep the largest distance 
$\tilde{d}_i$ for each point.
\item Sort the Mahalanobis distances in ascending order.
\item Among all $M$ trials, find the sequence $\tilde{d}_{[i]_M}$ 
with the minimum sum of distances from $n_{\epsilon}$ points.
\end{itemize}
\item Apply the expansion criteria to an increasing 
sequence of sets and determine
the region of interest for a structure.
\item In the region of interest find the 
largest estimate as $\hat\sigma$ and collect all points inside this scale.
\item Generate $N \ll M$ random trials from these points.
\begin{itemize}
\item Apply the mean shift to all the existing points,
to find the closest mode from $\alpha$.
\item Find $\hat\alpha$ at the maximum mode among all $N$ trials, 
and $\hat\btheta$ from the same elemental subset.
\item The recovered structure contains
$n_{in}$ points which converged to
$\pm \hat\sigma$ from $\hat\alpha$. 
\end{itemize}
\item Compute the TLS solution for the structure and
remove the $n_{in}$ points from the inputs. 
\item Go back to $\odot$ and start another iteration.
\item If not enough input points remain, 
sort all the structures by their strengths and return the result.
\end{itemize}
\noindent\hrulefill

\section{Experiments}
\label{sec:experiments}

Several synthetic and real examples are presented in this section.
In most cases a single carrier vector exists, $\zeta = 1$, 
except for the homography estimation which has two and $\zeta = 2$. 

The input data for synthetic problems are generated 
randomly and Gaussian noise is added to each inlier structure.
The standard deviation $\sigma_g$ is specified 
only to verify the results, while not used in the estimation process.

The real 3D datasets are constructed either from the 3D mesh models
in \cite{Remake} or the photogrammetric methods using 2D images.
Higher noise is introduced by the outliers which 
come from the incorrect point correspondences 
in both 2D and/or 3D matches.
In Fig.~\ref{figure_plane}a  a 2D image out of a  few tens of views of
a 2D image sequence is shown.
The image feature points are extracted from pairs of images and the point 
correspondences are robustly filtered through the above MISRE algorithm. 
The 3D information is recovered from the images 
based on the projective geometry relations.
As more images are registered, more 3D points are added and
the point cloud of the 3D scene is generated through incremental structure from 
motion (SfM)  \cite[Chapter 18]{hartley04} and  \cite{Remake, Wu13}, followed by
hierarchical merging \cite{Furu10, Farenzena09} (Fig.~\ref{figure_plane}b). 
See \cite[Chapter 5]{yang17} for a detailed description of the complete
recovery of 3D structures and \cite[Chapter 19]{hartley04} for more details of the auto-calibration processure.

The values of the scales and point amounts for each structure
are returned as the output of the algorithm.
The processing time on an i7-2617M 1.5GHz PC is also given.

\begin{figure}[t]
\centering
\begin{tabular}{@{\hspace{-0.2cm}}c@{\hspace{-0.0cm}}c}
\includegraphics[scale=0.3]{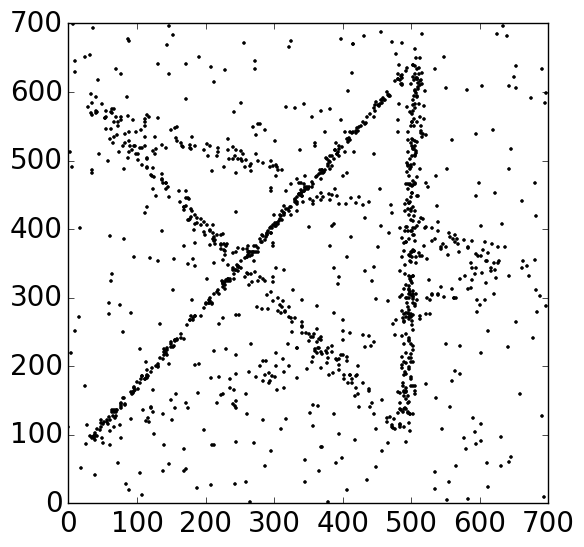}&
\includegraphics[scale=0.3]{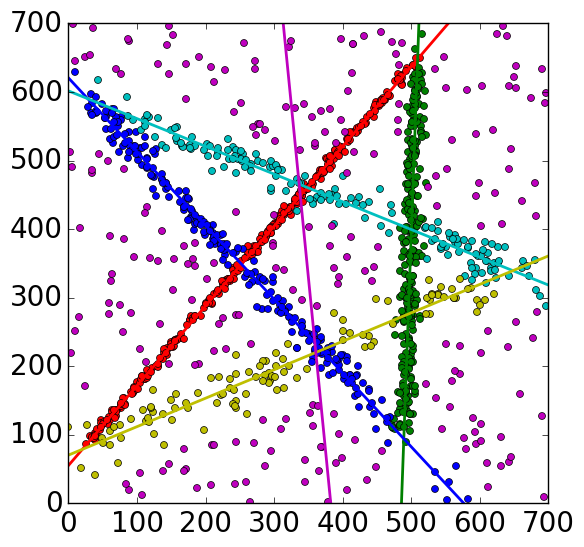} 
\end{tabular}
\begin{tabular}{@{\hspace{0.0cm}}c@{\hspace{0.1cm}}c}
(a) & (b) \\ 
\includegraphics[scale=0.2]{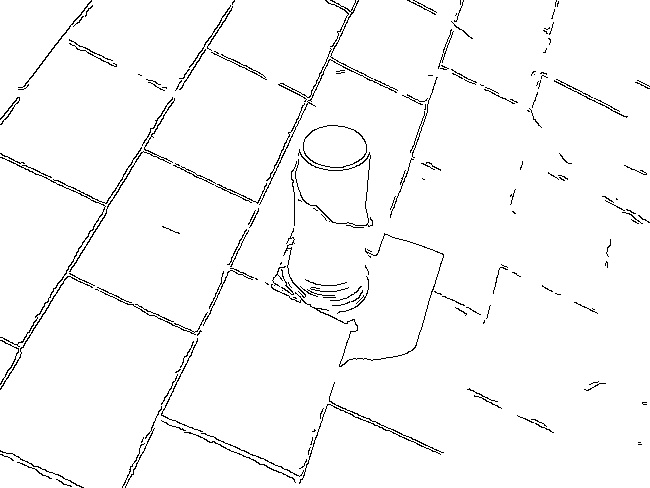}&
\includegraphics[scale=0.2]{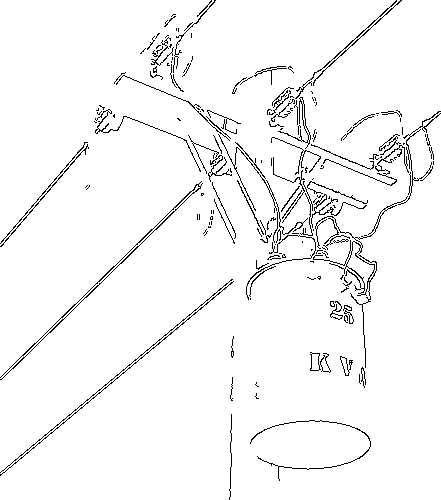}\\   
(c) & (d) \\
\includegraphics[scale=0.2]{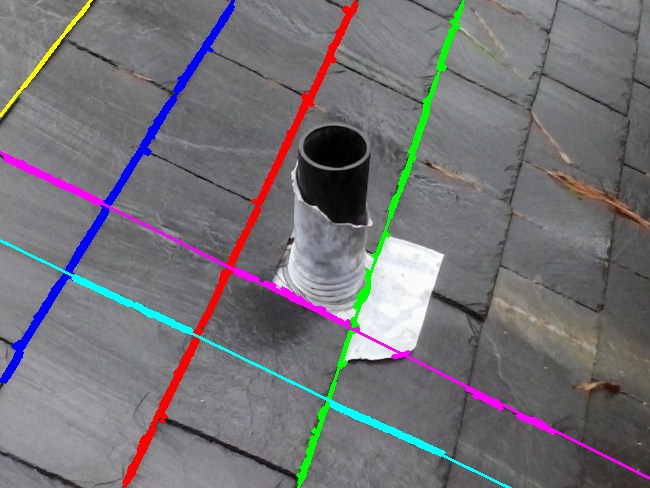}&
\includegraphics[scale=0.2]{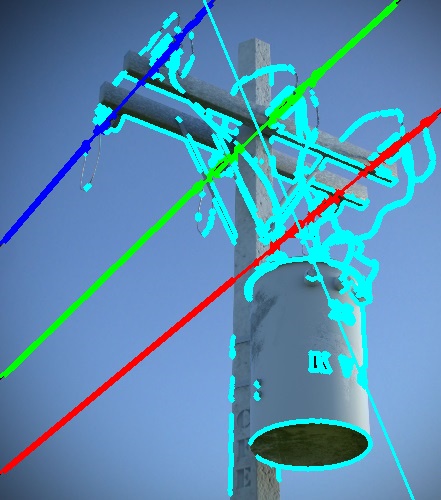}\\   
(e) & (f) 
\end{tabular}
\caption{2D lines estimations. 
(a) Synthetic Case: five lines with 350 outliers.  
(b) Recovered six structures. 
(c) {\it Roof}: Canny edges, 8310 points.
(d) {\it Pole}: Canny edges, 8072 points.
(e) {\it Roof}: Six strongest inlier structures.
(f) {\it Pole}: Three strongest inlier structures and one outlier structure.}
\label{fig:2dline}
\vspace{-.5cm}
\end{figure}

\subsection{2D Lines and 3D Planes}
\label{sec:2dline}

We first examine the use of MISRE in the estimation of linear 
geometric primitives.
For {\it multiple 2D lines}, the noisy objective function is
\begin{equation}
\label{eqn:line}
\theta_1 x_{i} + \theta_2 y_{i} - \alpha \simeq 0 \qquad 
i=1,\ldots,n_{in}.
\end{equation}
The input variable $\by = [x ~ y]^\top$ is identical with
the carrier vector $\bx$.

Five lines are placed in a $700\times700$
plane (Fig.\ref{fig:2dline}a) and
corrupted with different two-dimensional Gaussian noise.
They have $n_{in} = 300, 250, 200, 150, 100$ inlier points,
and $\sigma_g = 3, 6, 9, 12, 15$, respectively.
Another 350 unstructured
outliers are uniformly distributed in the image. 
The amount of points inside 
each inlier structure is 
small compared to the entire data.

With $M = 1000$, a test result is shown in Fig.\ref{fig:2dline}b. 
The algorithm recovers six structures
\[\begin{array}{rcccccc}
  & red  & green & blue & cyan & yellow & purple\\
scale: & 9.6  & 18.7 & 28.1 & 37.1 & 44.2 & 370.8\\
inliers: & 321 & 282 & 240 & 161 & 106 & 240\\
strength: & 33.4 & 15.1 & 8.5 & 4.3 & 2.4 & 0.6.
\end{array} \]
The first five structures are inliers with stronger strengths.
The sixth structure, is formed by outliers distributed over the whole image.

When the randomly generated inputs are tested 
independently for 100 times, 
the first four lines are correctly segmented in all the tests.
In the other six tests  
the weakest line ($n_{in}=100,~ \sigma_g = 15$) is not correctly
located.
Of  the 94 correct estimations,
the average result of the scale estimates and 
the classified inlier amounts as well as their respective  standard deviations are
\[\begin{array}{rccccc}
scale: & 10.48 & 19.94& 29.36 & 36.86 & 38.17\\
& (1.17) & (2.44) & (5.30) & (10.40) & (18.29)\\
inliers: & 335.9 & 285.8 & 240.8 & 155.6 & 93.4\\
& (8.2) & (9.5) & (21.3) & (27.0) & (28.4).\\
\end{array} \]
The average processing time is 0.58 seconds.
The estimated scale covers about  $3\sigma_g$ area of an inlier structure. 
In general, the number of classified inliers is larger than the
true amount due to the presence of outliers in the same area.

In Fig.\ref{fig:2dline}c and Fig.\ref{fig:2dline}d,
the Canny edge detection extracts 
similar sizes of input data (8310 and 8072 points)
from two real images. 
Again with $M=1000$,
the six strongest line structures are 
superimposed over the original image in Fig.\ref{fig:2dline}e.
In Fig.\ref{fig:2dline}f the three line structures together with the first 
outlier structure are shown.
The processing time depends on the number of structures that detected
by the estimator, 
these two estimations take 7.44  and 4.35 seconds, respectively.

For {\it multiple 3D planes},
the objective function for the inlier points is also linear
\begin{equation}
\label{eqn:line}
\theta_1 x_{i} + \theta_2 y_{i} + \theta_3 z_{i} - \alpha \simeq 0 \qquad 
i=1,\ldots,n_{in}.
\end{equation}
The input data are the 3D coordinates $\by = [x ~ y ~ z]^\top$ 
of the point cloud dataset.

\begin{figure}[t]
\centering
\begin{tabular}{@{\hspace{-0.0cm}}c@{\hspace{0.1cm}}c}
\includegraphics[scale=0.33]{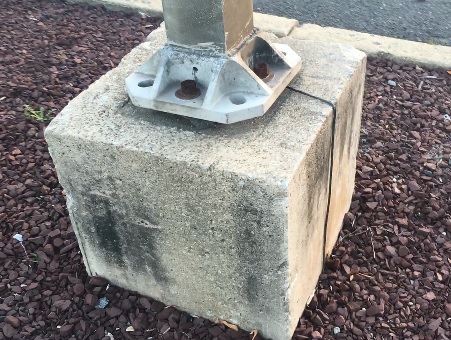}&
\includegraphics[scale=0.26]{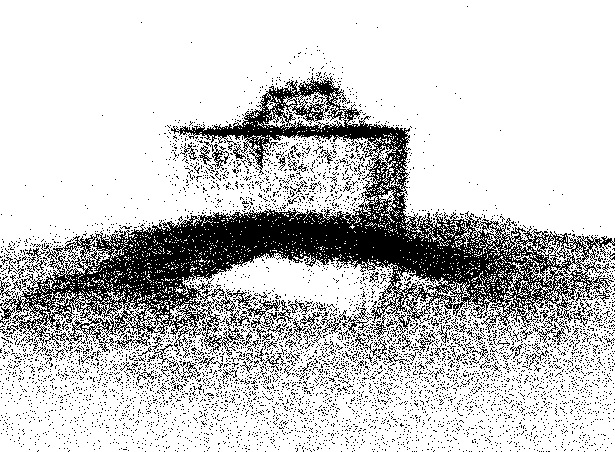} \\
(a) & (b) \\ 
\includegraphics[scale=0.33]{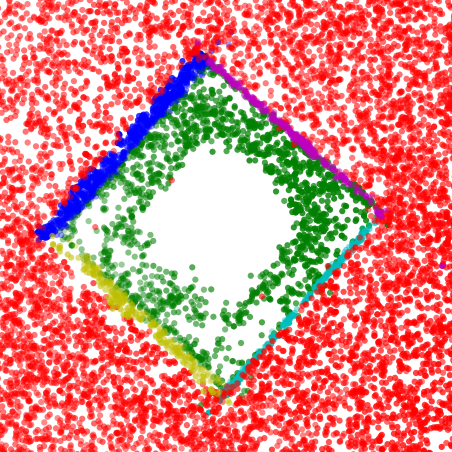}&
\includegraphics[scale=0.33]{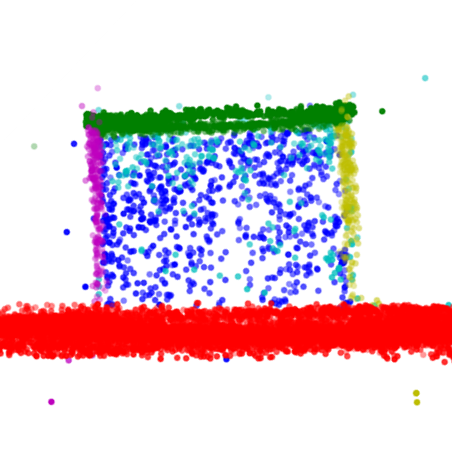} \\
(c) & (d)
\end{tabular}
\caption{3D Planes estimation in point cloud.
(a) A sample image used in SfM algorithm.
(b) A total of 23077 points selected.
(c) \& (d) Six planes recovered.}
\label{figure_plane}
\vspace{-.5cm}
\end{figure}

In Fig.\ref{figure_plane}a we show a sample image used in 
the SfM algorithm \cite{Wu13, Furu10},
and obtain the point cloud as in Fig.\ref{figure_plane}b. 
The total of 23077 points also with outliers, are recovered from 70 
2D images.
After 7.04 seconds, with $M = 1000$,
the estimator locates six planes as shown in Fig.\ref{figure_plane}c
and Fig.\ref{figure_plane}d with 21758 inlier points in total.

\subsection{2D Ellipses}
\label{sec:2dellipse}

In the next experiment {\it multiple 2D ellipses} are estimated. 
The noisy objective function is
\begin{equation}
\label{eqn:ellipse}
(\by_i - \by_c)^\top \bQ (\by_i - \by_c) - 1 \simeq 0
\qquad i=1,\ldots,n_{in}
\end{equation}
where $\bQ$ is a symmetric $2\times 2$ positive definite matrix and
$\by_c$ is the position of the ellipse center. Given the input variable 
$\by = [x ~ y]^\top$, the carrier is derived as 
$\bx = [x ~ y ~ x^2 ~ xy ~ y^2 ]$. 
The condition $4 \theta_3 \theta_5 - \theta_4^2 > 0$ also
has to be satisfied in order to represent an ellipse.
We also enforce the constraint that 
the major axis cannot be more than 10 times longer than the 
minor axis, to avoid classifying line segment as
a part of a very flat ellipse.

The transpose of the $5\times 2$ Jacobian matrix is
\begin{equation}
\label{eqn:ellipjacob}
\bJ_{\scriptsize{\mathbf{x} | \mathbf{y}}}^\top = 
\left[ \begin{array}{ccccc} 
1 & 0 & 2x & y & 0 \\
0 & 1 & 0  & x & 2y
\end{array} \right] .
\end{equation}
The ellipse fitting is a nonlinear estimation and biased,
especially for the part with large curvature. 
When the inputs are perturbed with zero mean Gaussian noise with 
$\sigma_g$, the standard deviation of carrier vector $\bx$
relative to the true value $\bx_o$ is not zero mean
\begin{equation}
\bE (\bx -\bx_o) = [0 ~~~ 0 ~~~ \sigma^2_g ~~~ 0 ~~~ \sigma^2_g ]^\top
\end{equation}
since the carrier contains $x^2,\;y^2$ terms.
A bias in the estimate can be clearly seen 
when only a small segment of the noisy ellipse is given
in the input. 
Taking into account also the second order statistics 
in estimation still does not  eliminate the bias. 
See  papers \cite{kanatani06},
\cite{szpak15} and their references for additional methods.

\begin{figure}[t]
\centering
\begin{tabular}{@{\hspace{-0.0cm}}c@{\hspace{0.0cm}}c}
\includegraphics[scale=0.3]{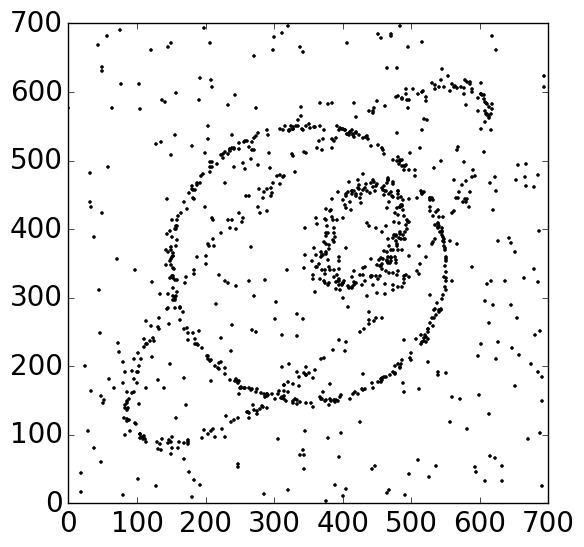}&
\includegraphics[scale=0.3]{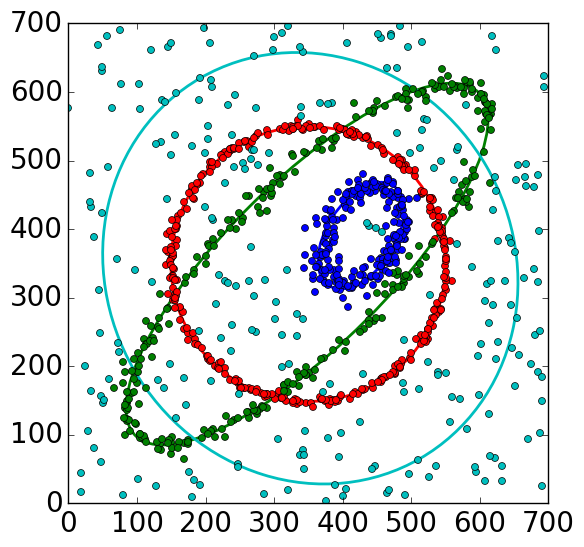}\\   
(a) & (b) \\
\includegraphics[scale=0.3]{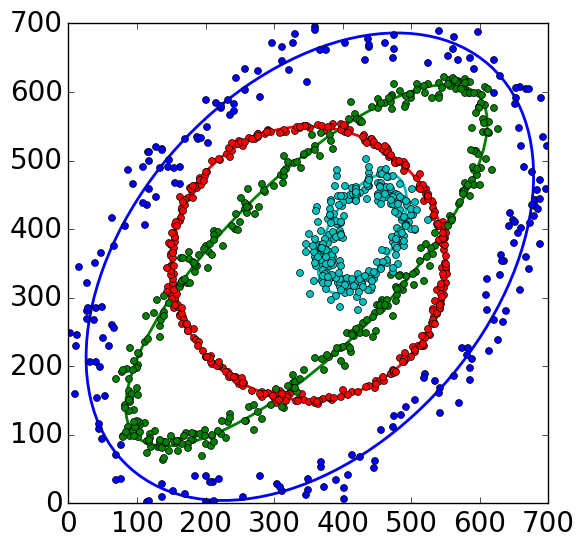}&
\includegraphics[scale=0.3]{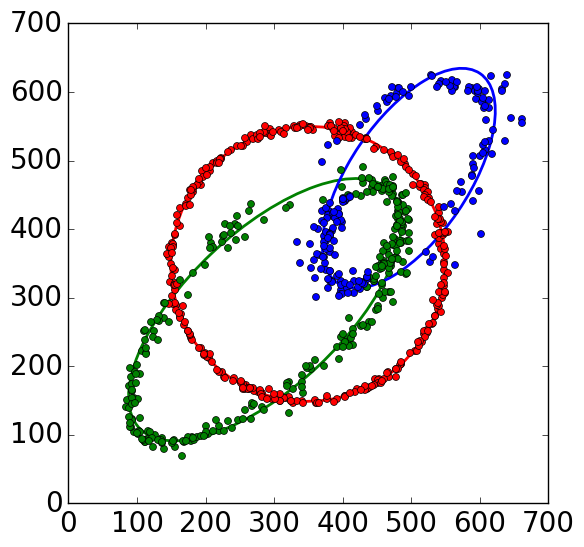}\\   
(c) & (d)
\end{tabular}
\caption{Synthetic 2D ellipse estimations. 
(a) Case 1: three ellipses and 350 outliers.
(b) Recovered four structures.
(c) Case 2: outliers increased to 800, recovered four structures.
(d) Interaction between two ellipses.}
\label{fig:synthellipse}
\vspace{-.5cm}
\end{figure}

\begin{figure}[t]
\centering
\begin{tabular}{@{\hspace{-0.0cm}}c@{\hspace{0.1cm}}c}
\includegraphics[scale=0.17]{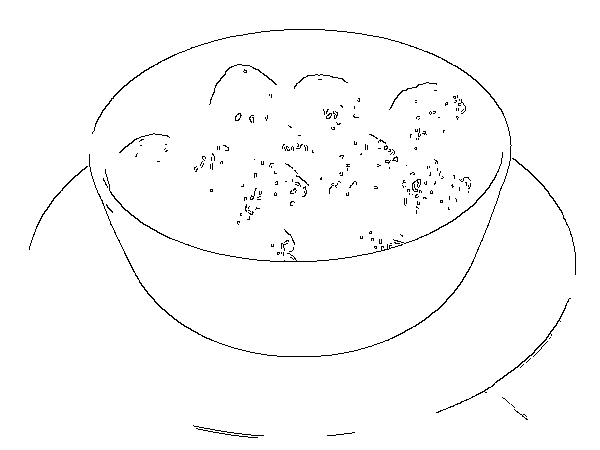}&
\includegraphics[scale=0.3]{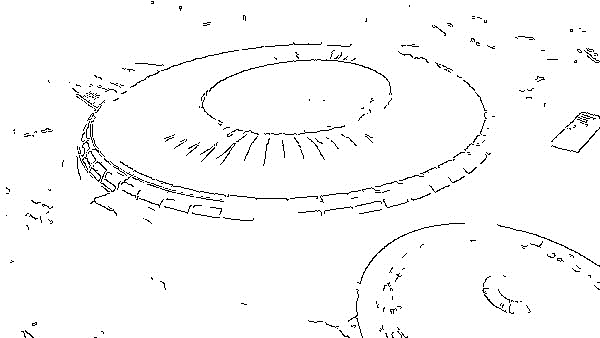}\\   
(a) & (b) \\
\includegraphics[scale=0.17]{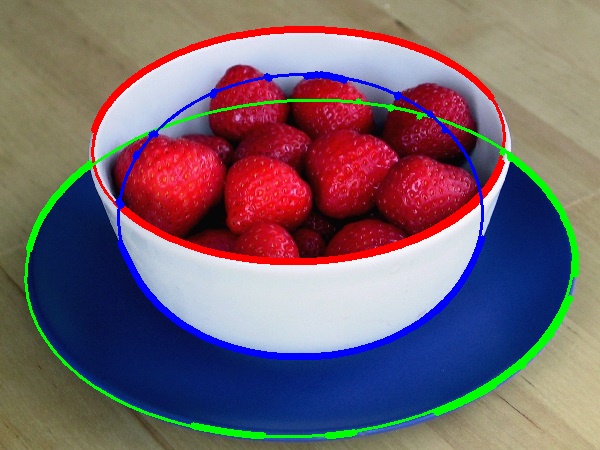}&
\includegraphics[scale=0.3]{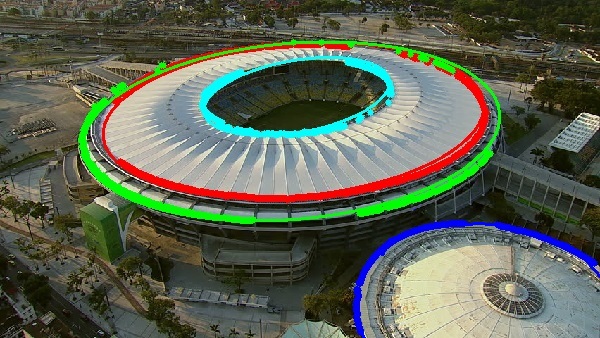}\\   
(c) & (d) 
\end{tabular}
\caption{2D ellipses in real images. 
(a) {\it Strawberries}: Canny edges, 4343 points.
(b) {\it Stadium}: Canny edges, 4579 points.
(c) {\it Strawberries}: Three strongest inlier structures.
(d) {\it Stadium}: Four strongest inlier structures (see also text).}
\label{fig:realellipse}
\vspace{-.5cm}
\end{figure}

In Fig.\ref{fig:synthellipse}a three ellipses are placed with 350 outliers
in the background. The inlier structures have
$n_{in} = 300, 250, 200$ and $\sigma_g = 3, 6, 9$.
The smallest ellipse with $n_{in} = 200$ is corrupted with the largest noise $\sigma_g = 9$.
We use $M = 5000$ in the ellipse fitting experiments.
When tested (Fig.\ref{fig:synthellipse}b), four ellipses are recovered 
\[\begin{array}{rcccc}
  & red  & green & blue & cyan\\
scale: & 12.1  & 28.9 & 48.0 & 1321.2\\
inliers: & 337 & 292 & 222 & 248\\
strength: & 28.0 & 10.1 & 4.6 & 0.2.
\end{array} \]
Based on results sorted by strength, 
the first three structures are inliers and are returned first.

When the estimation is repeated 100 times, the three inlier 
structures are correctly located 97 times,
while in the other three tests 
the smallest ellipse is not estimated correctly.
From the 97 correct estimations, 
the average scales, the classified inlier amounts, 
along with their standard deviations are
\[\begin{array}{rccc}
scale: & 11.60 & 21.59 & 32.87 \\
& (1.54) & (3.64) & (14.71) \\
inliers: & 336.2 & 272.9 & 196.4 \\
& (8.2) & (27.5) & (53.2). \\
\end{array} \]
The average processing time is 3.28 seconds.

When the outlier amount reaches the limit,
the inlier structure with weakest strength may
no longer be sorted before the outliers.
The scale estimate becomes inaccurate due to the 
heavy outlier noise, and the outliers can form more dense 
structure with comparable strength.
When 800 outliers 
are placed in the image,
a test gives the result in Fig.\ref{fig:synthellipse}c.
The outlier structure (blue) has a strength of 4.8, while the value of 
the inliers (cyan) is 3.9.
However, the first two inlier structures are still recovered 
due to their stronger strengths. 

We also gives an example to show 
one of the limitation explained in Section \ref{sec:inlieroutlier},
when the inlier strength is too weak to tolerate more outliers.
In Fig.\ref{fig:synthellipse}d two inlier structures interact, 
and the mean shifts converge to incorrect modes.

From Canny edge detection, 4343 and 4579 points are obtained 
from two real images containing several objects with elliptic shapes,
as shown in Fig.\ref{fig:realellipse}a and Fig.\ref{fig:realellipse}b.
With $M=5000$, the three strongest ellipses are drawn in 
Fig.\ref{fig:realellipse}c, superimposed over the original images.
The processing time is 18.90 seconds in this case.
In Fig.\ref{fig:realellipse}d the estimation takes 23.14 seconds to detect
four strongest ellipses, which are inlier structures. 
After 100 repetitive tests using the data shown 
in Fig.\ref{fig:realellipse}b, 
only the first two ellipses (red and green)
are detected reliably in 98 times.
The other two ellipses (blue and cyan) have smaller amounts of inliers
and therefore are less stable.
The data acquired from Canny edge detection
do not necessarily render the overall inlier structures in 
a more dense state. 
Preprocessing on the edge data is generally required for better performance.

\subsection{3D Spheres}
\label{sec:3dsphere}

For spherical surface fitting in point cloud,
the objective function of is
\begin{equation}
\label{eqn:sphere}
(x - a)^2 + (y - b)^2 + (z - c)^2 - r^2 \simeq 0 
\end{equation}
a carrier vector $\bx \in \mathbb{R}^4$ is derived with $m=4$ carriers
$\bx=\left[x^2 + y^2 + z^2\;\;x\;\;y\;\;z \right]^\top$ .
The transpose of the $4\times 3$ Jacobian matrix
of a spherical surface is derived as
\begin{equation}
\label{eqn:sphereJac}
\bJ_{\scriptsize{\mathbf{x} | \mathbf{y}}}^\top = \left[
\begin{array}{cccc}
2x_i & 1 & 0 & 0\\ 
2y_i & 0 & 1 & 0\\ 
2z_i & 0 & 0 & 1
\end{array} \right].
\end{equation}

\begin{figure}[t]
\centering
\begin{tabular}{@{\hspace{-0.0cm}}c@{\hspace{0.2cm}}c}
\includegraphics[scale=0.23]{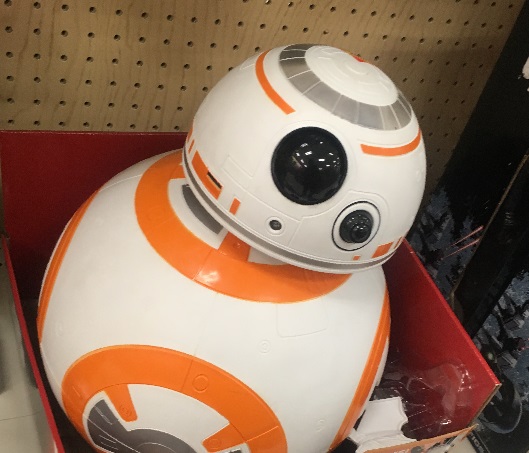}&
\includegraphics[scale=0.33]{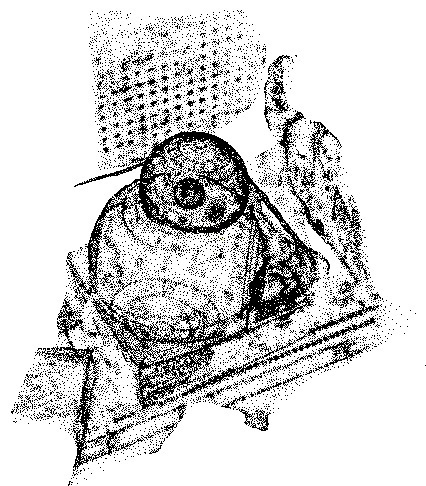} \\
(a) & (b) \\
\includegraphics[scale=0.26]{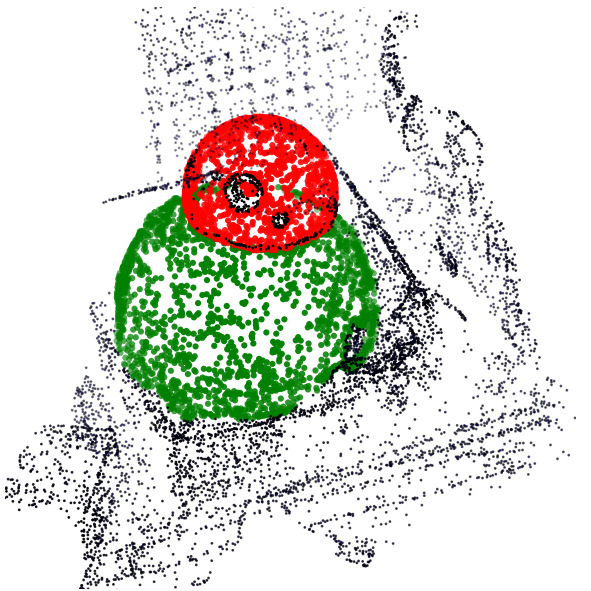}&
\includegraphics[scale=0.26]{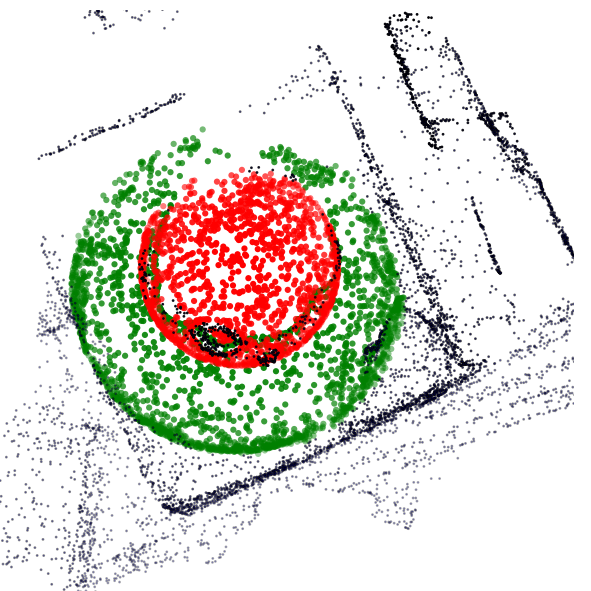}\\   
(c) & (d)
\end{tabular}
\caption{3D spheres estimation in point cloud.
(a) An image from the sequence.
(b) A total of 10854 points selected.
(c) \& (d) Two spheres recovered.
}
\label{figure_7}
\vspace{-.5cm}
\end{figure}

A sample image is shown in Fig.\ref{figure_7}a where
a toy with spherical surfaces is portrayed.
In Fig.\ref{figure_7}b the 3D point cloud is generated 
with \cite{Remake}, containing 10854 points from 36 images in 2D.
A large number of points in the background have to be 
rejected as outliers.
With $M = 1000$, we locate two inlier structures,
as shown in red and green colors in Fig.\ref{figure_7}c
and Fig.\ref{figure_7}d. 
A total of 3504 points are inliers and the estimation took 7.24 seconds.

\subsection{3D Cylinders}
\label{sec:3dcylinder}

\begin{figure}[t]
\centering
\begin{tabular}{@{\hspace{-0.0cm}}c@{\hspace{-0.2cm}}c}
\includegraphics[scale=0.33]{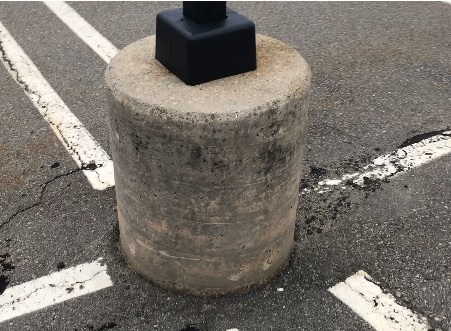}&
\includegraphics[scale=0.24]{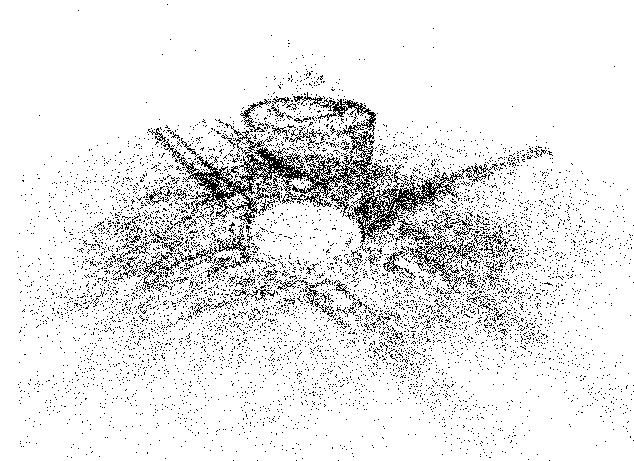} \\
(a) & (b) \\
\includegraphics[scale=0.3]{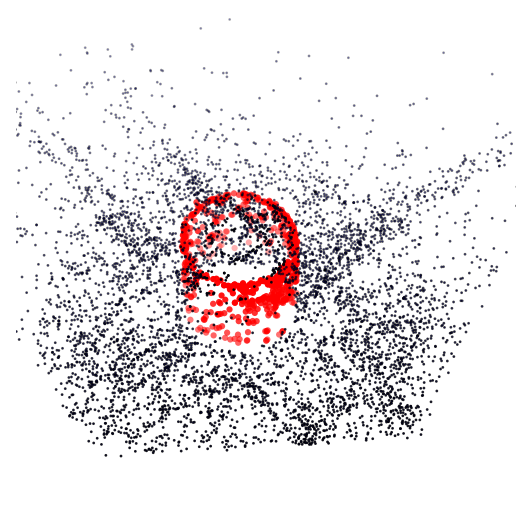}&
\includegraphics[scale=0.3]{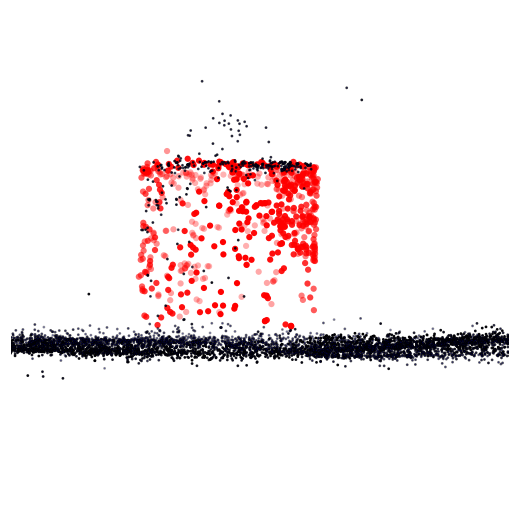} \\
(c) & (d) \\
\includegraphics[scale=0.15]{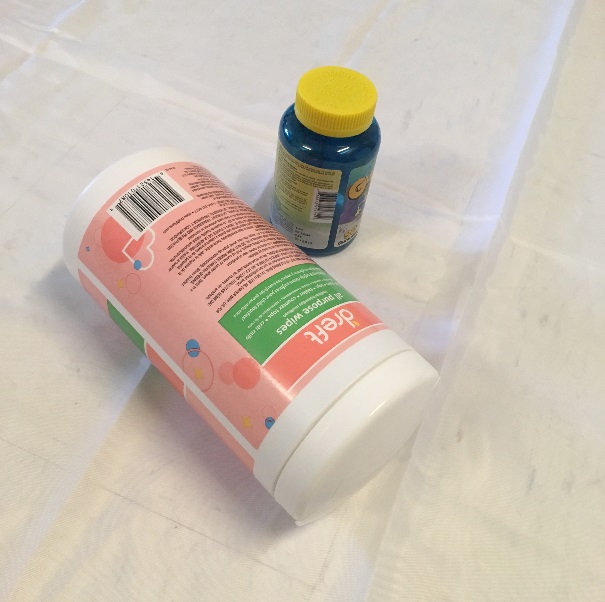}&
\includegraphics[scale=0.3]{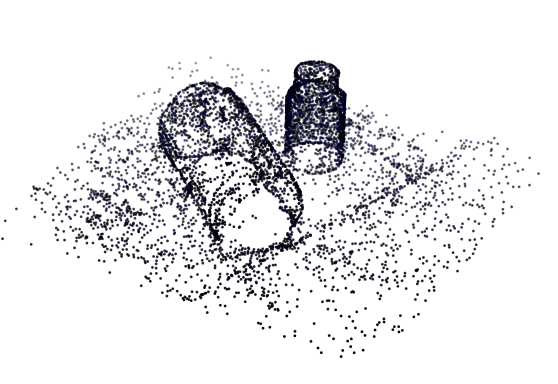} \\
(e) & (f) \\
\includegraphics[scale=0.3]{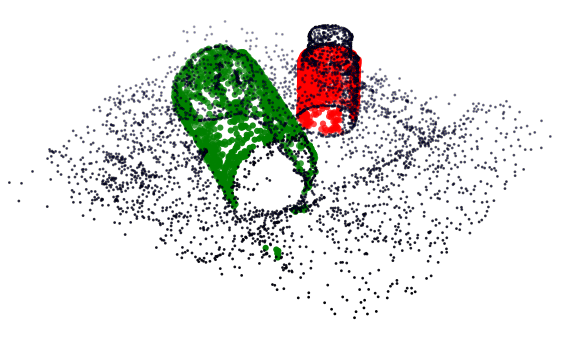}&
\includegraphics[scale=0.25]{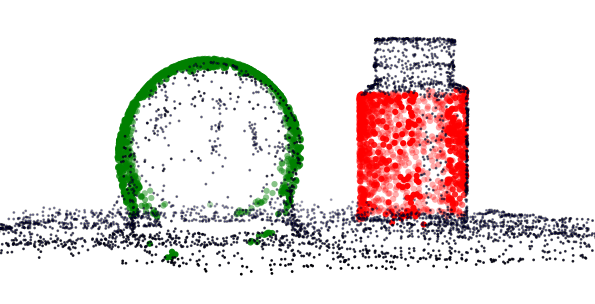} \\
(g) & (h)
\end{tabular}
\caption{3D cylinders estimation in point cloud.
(a) A sample image containing one cylinder.
(b) A total of 7241 points selected.
(c) \& (d) One cylinder recovered.
(e) A sample image containing two cylinders.
(f) A total of 6500 points selected.
(g) \& (h) Two cylinders recovered.
}
\label{figure_9}
\vspace{-0.5cm}
\end{figure}

A {\it cylinder} aligned with the Z-axis is defined by the equation
\begin{equation}
(x - a)^2 + (y - b)^2 - r^2 = 0 
\end{equation}
$a,~b$ stand for the 2D coordinates where Z-axis
passes through the XY-plane, and $r$ is the radius.
With the input variable $\by = [x ~ y ~ z]^\top$,
this relation can be reformulated by a quadric matrix
$[\by_i ~ 1]\, \bP' \, [\by_i ~ 1]^\top \simeq 0
~~ i=1,\ldots,n_{in}$
where $\bP'$ is a $4\times 4$ symmetric matrix
\begin{eqnarray}
\label{eqn:cylinderQuadric}
\nonumber
\bP' = \lambda \begin{bmatrix} \bD' & \bd' \\ \bd^{'T} & a^2 + b^2 -r^2 \end{bmatrix}
\\  \bD' = \begin{bmatrix} 1 & 0 & 0\\ 0 & 1 & 0 \\ 0 & 0 & 0 \end{bmatrix}
\ \  \bd' = \begin{bmatrix} -a \\ -b \\ 0 \end{bmatrix}
\end{eqnarray}
when an euclidean transformation is applied
$\bM = \begin{bmatrix} \bR & \mathbf{t} \\ \mathbf{0}^{T} & 1 \end{bmatrix}$, 
a general cylinder under rotation and translation is found with
\begin{eqnarray}
\label{eqn:cylinderP}
\bP = \bM^{-T} \bP' \bM^{-1} = 
\begin{bmatrix} \bD & \bd \\ \bd^{T} & d \end{bmatrix}.
\end{eqnarray}
The $4\times 4$ quadric matrix $\bP$ has nine unknown 
parameters up to a scale.
A general cylinder have only five degrees of freedom, 
four for its axis of rotation and one for radius. 

Several solutions of the cylinders were described in \cite{Beder06}, 
computed at elemental subsets with various numbers of points
from 5 to 9.
The nine-point-solution is used in this experiment,
where the carrier vector is derived as
$\bx = [x^2 ~~ xy ~~ xz ~~ y^2 ~~ yz ~~ z^2 ~~ x ~~ y ~~ z]$. 
An elemental subset consisting of nine points gives an over-determined solution, and the parameters in $\btheta$ should be constrained 
for a cylinder.
From equations (\ref{eqn:cylinderQuadric}) and 
(\ref{eqn:cylinderP}), 
it is easy to prove that two of the three singular values of matrix $\bD$ are identical and the third one is zero,
and $\bd$ is an eigenvector of $\bD$.
These constraints should be verified for each elemental subset.

The transpose of the $9\times 3$ Jacobian matrix is
\begin{equation}
\label{eqn:cylinderjacob}
\bJ_{\scriptsize{\mathbf{x}_i | \mathbf{y}_i}}^\top = 
\left[ \begin{array}{ccccccccc} 
2x_i & y_i & z_i & 0 & 0 & 0 & 1 & 0 & 0\\
0 & x_i & 0 & 2y_i & z_i & 0 & 0 & 1 & 0\\
0 & 0 & x_i & 0 & y_i & 2z_i & 0 & 0 & 1
\end{array} \right] .
\end{equation}

A cylindrical pole is shown in Fig.\ref{figure_9}a.
From 54 images we generate the point cloud
(Fig.\ref{figure_9}b) through the
structure from motion (SfM) algorithm.
The dataset contains 7241 points with most on the ground
being outliers for cylinder detection.
With $M = 2000$, after 18.55 seconds the single 
cylinder is located in the noisy dataset containing 568 points, 
as the red points shown
in Fig.\ref{figure_9}c and Fig.\ref{figure_9}d.

In Fig.\ref{figure_9}e another sample image is shown.
A total of 6500 vertices are obtained in Fig.\ref{figure_9}f
from the mesh rebuilt in \cite{Remake}, by using 22 images.
The two bottles with cylindrical shapes are detected
in 12.65 seconds.
The two inlier structures are shown in Fig.\ref{figure_9}g and Fig.\ref{figure_9}h containing 2262 points.

\vspace{-.1cm}
\subsection{Fundamental Matrices}
\label{sec:fundmat}

The next experiment shows the estimation of the {\it fundamental matrices}. 
The corresponding linear space was introduced in Section \ref{sec:transformation}. 
The $3\times 3$ matrix $\bF$ is rank-2 and 
a recent paper \cite{cheng15} solved the 
non-convex problem iteratively 
by a convex moments based polynomial optimization.
It compared the results with a few RANSAC type algorithms.
The method required several parameters and
in each example only one fundamental matrix was recovered.

\begin{figure}[t]
\centering
\begin{tabular}{@{\hspace{-0.0cm}}c@{\hspace{-0.0cm}}c}
\includegraphics[scale=0.3125]{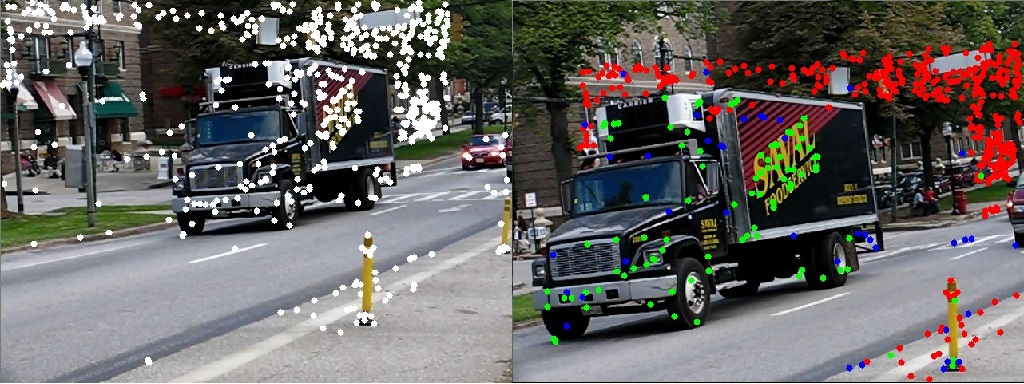}\\
(a) \\ 
\includegraphics[scale=0.3125]{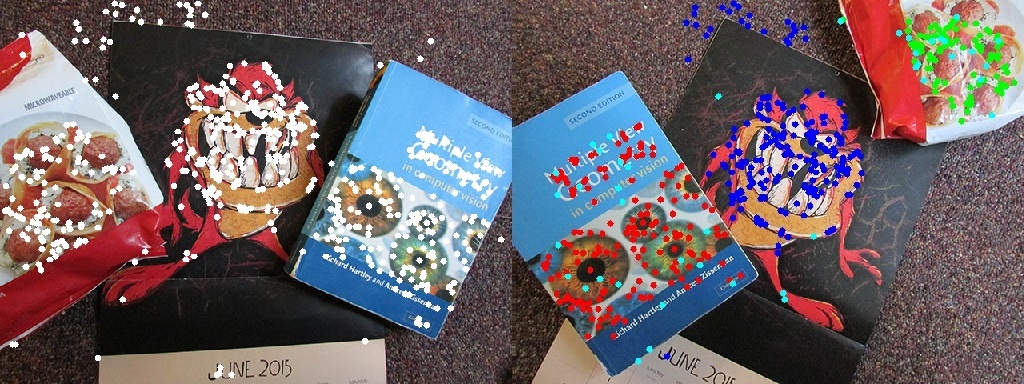}\\   
(b) \\
\includegraphics[scale=0.3125]{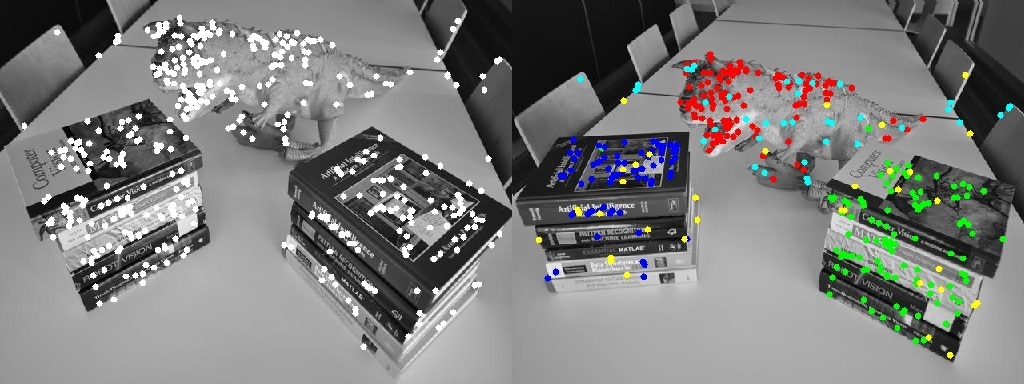}\\   
(c)
\end{tabular}
\caption{Fundamental matrices estimation. 
(a) Image pair from {\it Hopkins 155 dataset} with two inlier
and one outlier structures.
(b) The {\it books} with three inlier and one outlier structures.
(c) The {\it dinabooks} from \cite{pham14} with 
four inlier and one outlier structures.}
\label{fig:fundmatFig2}
\vspace{-.5cm}
\end{figure}

The fundamental matrix cannot be used directly to segment objects 
with only translational motions, as proved in \cite{babHadiashar09}.
Each example of Fig.\ref{fig:fundmatFig2} shows the movement of multiple rigid objects, where a large enough rotation exists.
The point correspondences are extracted by
OpenCV with a distance ratio of 0.8 for SIFT \cite{lowe04},
giving 608, 614 and 457 matches, respectively.
With $M=5000$, the structures are retained as
\[\begin{array}{rccc}
\mbox{Fig.\ref{fig:fundmatFig2}a}  & red  & green & blue\\
scale: & 0.56  & 0.73 & 11.78\\
inliers: & 407 & 101 & 51\\
strength: & 727.3 & 139.3 & 4.33.
\end{array} \]
\[\begin{array}{rcccc}
\mbox{Fig.\ref{fig:fundmatFig2}b}  & red  & green & blue & cyan\\
scale: & 0.46  & 0.40 & 1.12 & 10.42\\
inliers: & 192 & 96 & 221 & 47\\
strength: & 413.4 & 242.8 & 196.8 & 4.5.
\end{array} \]
\[ \begin{array}{rccccc}
\mbox{Fig.\ref{fig:fundmatFig2}c}  & red  & green & blue & cyan & yellow\\
scale: & 0.22  & 0.72 & 0.65 & 0.70 & 23.9\\
inliers: & 135 & 117 & 84 & 48 & 43\\
strength: & 623.0 & 161.5 & 129.0 & 68.2 & 1.8.
\end{array} \]
The estimations take 1.75, 2.30 and 2.10 seconds for these three cases.
In real images, the outlier structures can be easily filtered out since they
have much larger scales than the inliers. 
It can be observed that the scales of the inlier structures are very close,
therefore the methods with fixed thresholds may be used here. 
However, if the images are scaled before estimation, the error in the inliers will change proportionally. Correct scale estimate can only be found adaptively from the input data.

As discussed in Section \ref{sec:inlieroutlier},
the first (red) and the fourth (cyan) structures obtained
from Fig.\ref{fig:fundmatFig2}c can be fused as a single structure.
This merge has to be done by post-processing in the input space but
also requires a threshold from the user.

In SIFT matches false correspondences always exist.
If the images contain repetitive features, 
such as the exterior of buildings,
parametrization of the repetitions can reduce the uncertainty
\cite{Schaffalitzky99}. 
Preprocessing of the images is not described 
in this paper, and we will not explain it further.

\subsection{Homographies}
\label{sec:2dhomography}

\begin{figure*}[t]
\centering
\begin{tabular}{cc}
\includegraphics[scale=0.21]{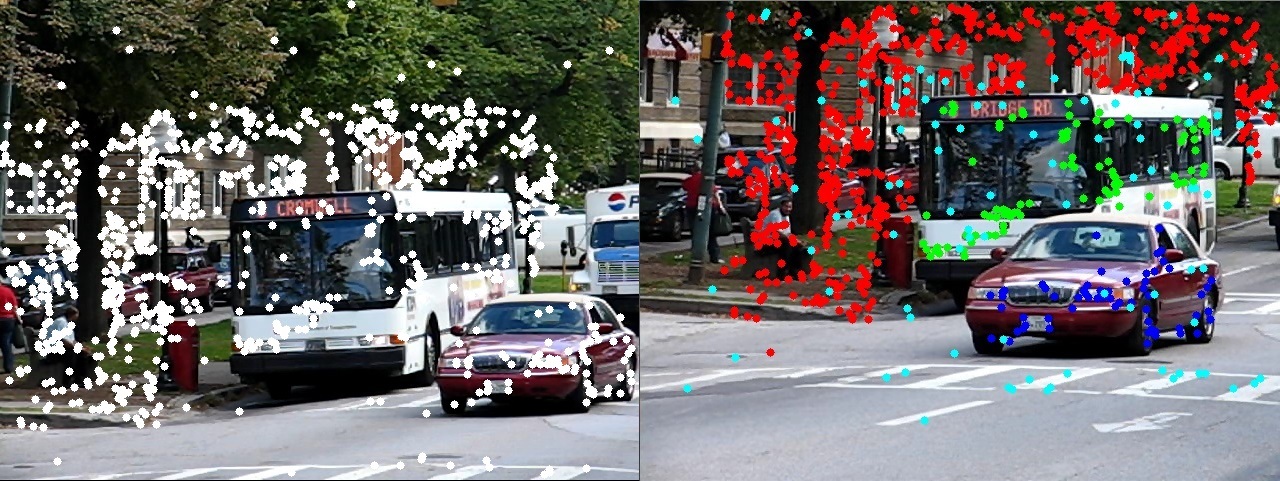}&
\includegraphics[scale=0.42]{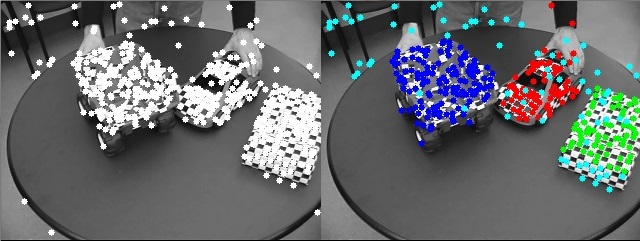}\\
(a) & (b)\\
\includegraphics[scale=0.21]{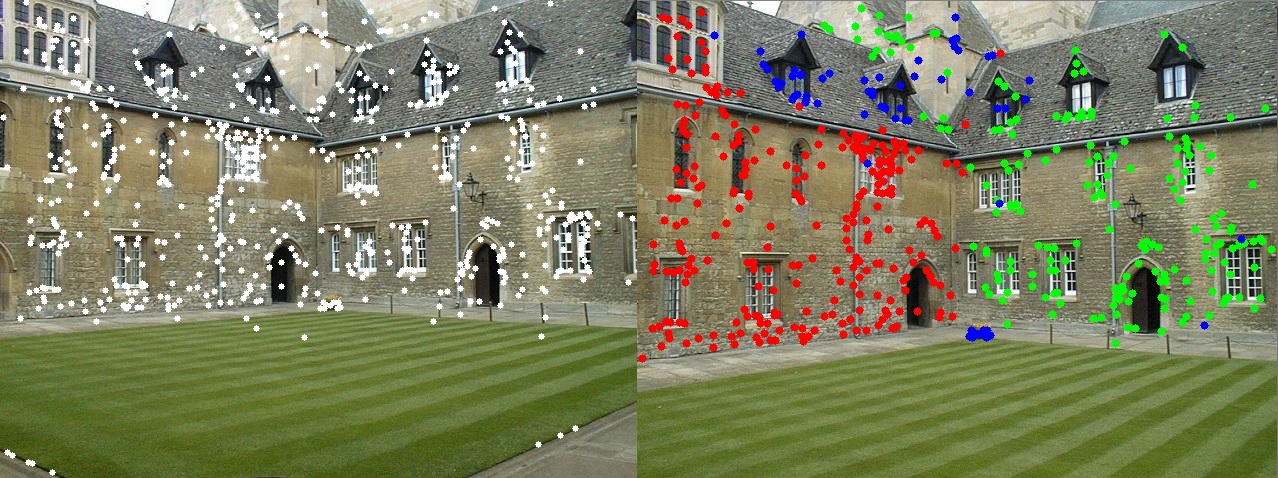}&
\includegraphics[scale=0.22]{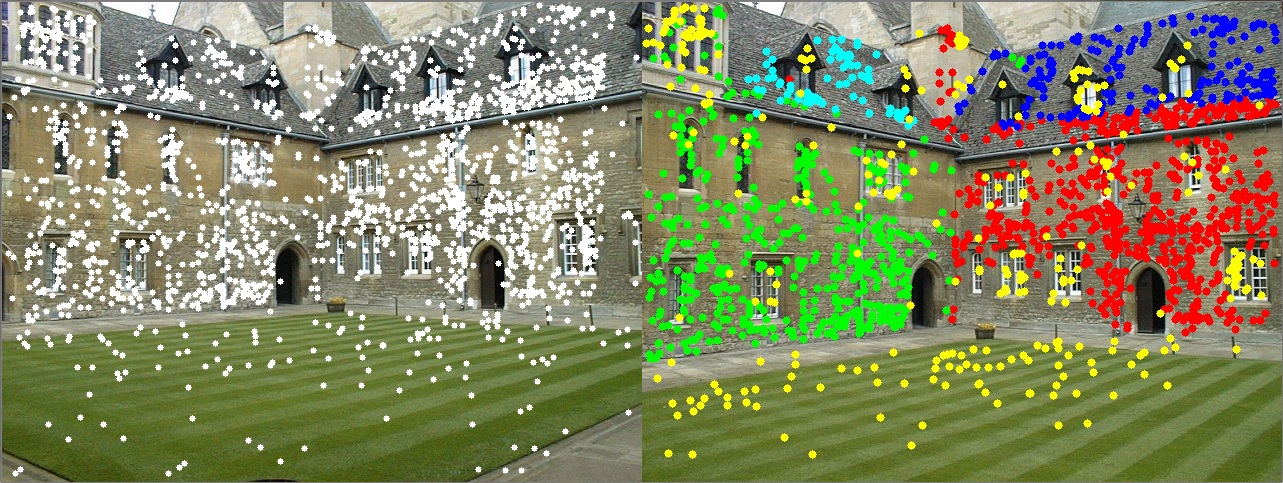}\\
(c) & (d) 
\end{tabular}
\caption{Homography estimation.
(a) {\it Hopkins 155}: Three inlier and one outlier structures. 
(b) {\it Hopkins 155}: Three inlier and one outlier structures.
(c) {\it Merton College} with 713 point pairs.
(d) {\it Merton College} with 1940 point pairs.}
\label{fig:homographyFig1}
\vspace{-.5cm}
\end{figure*}

The final example is for 2D {\it homography} estimation.
Each inlier structure is represented by a $3\times 3$ matrix $\bH$, 
which connects two planes inside the image pair
\begin{equation}
\label{eqn:homography}
\mathbf{y}_{i}^\prime \simeq \bH \by_i,\quad i = 1,\ldots,n_{in}
\end{equation}
where $\by = [x ~ y ~ 1]^\top$ and 
$\by^\prime = [x^\prime ~ y^\prime ~ 1]^\top$ are the homogeneous
coordinates in these two images. 

As mentioned in Section \ref{sec:prepare}, 
the homography estimation has $\zeta =2$. 
The input variables are $[x~~y~~x^\prime~~y^\prime]^\top$.
Two linearized relations can be derived from the
constraint (\ref{eqn:homography}) by the direct linear transformation (DLT)
\begin{equation}
\label{eqn:DLThomo}
\mathbf{A}_i\mathbf{h} = \left[
\begin{array}{@{\hspace{-0.03cm}}c@{\hspace{0.1cm}}c@{\hspace{0.1cm}}c@
{\hspace{-0.00cm}}}
-\mathbf{y}_i^{\top} & \phantom{-}\mathbf{0}^{\top}_3 &
\phantom{-}x_{i}^{\prime}\mathbf{y}_i^{\top} \\
\mathbf{0}^{\top}_3  & -\mathbf{y}_i^{\top} &
\phantom{-}y_{i}^{\prime}\mathbf{y}_i^{\top} \\
\end{array}\right]
\left[\begin{array}{@{\hspace{-0.03cm}}c@{\hspace{-0.03cm}}}
\mathbf{h}_1\\ \mathbf{h}_2\\ \mathbf{h}_3\\
\end{array} \right]
\simeq \mathbf{0}_2~.
\end{equation} 
The matrix $\bA_i$ is $2\times 9$ and both rows satisfy 
the relations with the vector derived from the matrix 
vec$(\bH^\top) = \bh = \btheta$.

The carriers are obtained from the two rows of $\bA_i$
\begin{eqnarray}
\label{eqn:carriershomography}
\bx^{[1]} &=& [ {-x} ~~ {-y} ~~ {-1} ~~ 0 ~~ 0 ~~ 0 ~~ x^\prime x ~~ 
x^\prime y ~~ x^\prime ]^\top \nonumber \\
\bx^{[2]} &=& [ 0 ~~ 0 ~~ 0 ~~  {-x} ~~ {-y} ~~ {-1} ~~ y^\prime x ~~ 
y^\prime y ~~ y^\prime ]^\top .
\end{eqnarray}
The transpose of the two $9\times 4$ Jacobians matrices are
\begin{align}
\nonumber\mathbf{J}_{\scriptsize{\mathbf{x}_i^{[1]}|\mathbf{y}}}^\top &=
\left[
\begin{tabular}{@{\hspace{-0.01cm}}c@{\hspace{-0.01cm}}c@
{\hspace{-0.001cm}}c@{\hspace{-0.01cm}}} {$-\mathbf{I}_{2\times 2}$} &
\multirow{3}{*}{$\phantom{-}\mathbf{0}_{4\times 4}$} &
$\phantom{-}x^\prime_i\mathbf{I}_{2\times 2}~~\mathbf{0}_2$\\
{$\phantom{0}\mathbf{0}_{2}^\top$} & & $\phantom{--}\mathbf{y}_i^\top$\\
{$\phantom{0}\mathbf{0}_{2}^\top$} & &
{$\phantom{--}\mathbf{0}_{2}^\top~~~~0$}\\
\end{tabular} \right]\\
\mathbf{J}_{\scriptsize{\mathbf{x}_i^{[2]}|\mathbf{y}}}^\top &=
\left[\begin{tabular}{@{\hspace{-0.01cm}}ccc@{\hspace{-0.01cm}}c@
{\hspace{-0.01cm}}} \multirow{3}{*}{$\mathbf{0}_{4\times 3}$} &
$-\mathbf{I}_{2\times 2}$ & \multirow{3}{*}{$\mathbf{0}_{4}$} &
{$\phantom{0}y^\prime_i\mathbf{I}_{2\times 2}~~~\mathbf{0}_2$} \\
& {$\mathbf{0}_{2}^\top$} & & $\phantom{-}\mathbf{0}_{2}^\top~~~~~0$ \\
& {$\mathbf{0}_{2}^\top$} & & $\phantom{--}\mathbf{y}_i^\top$ \\
\end{tabular}
\right] .
\end{align}
Based on Section \ref{sec:reducedistance}, for every $\btheta$
only the larger Mahalanobis distance is used for each
input $\by_i, \; i=1,\ldots,n$.

The motion segmentation involves only small translation in 3D in
Fig.\ref{fig:homographyFig1}a and Fig.\ref{fig:homographyFig1}b,
both images are taken from the {\it Hopkins 155} dataset.
With $M=2000$, 
the processing time is 1.12 and 1.09 seconds for the inputs containing 990 and 482 SIFT point pairs, respectively.
As mentioned in Section \ref{sec:inlieroutlier}, in 
Fig.\ref{fig:homographyFig1}a 
the estimator cannot separate these two 3D planes on the bus
because the 2D homographies corresponding to them are very similar.
The condition (\ref{eqn:condition}) does not stop where it should 
since the points on either 2D planes are not dense enough.
The example shows that the desired results can only be obtained by increasing the amount of inliers from preprocessing.
In Fig.\ref{fig:homographyFig1}b, the three objects can
be correctly separated in spite of the very small motions, 
due to the stronger strengths in all the inlier structures.

\[\begin{array}{rcccc}
\mbox{Fig.\ref{fig:homographyFig1}a}  & red  & green & blue & cyan\\
scale: & 1.62  & 1.12 & 4.49 & 203.11\\
inliers: & 713 & 101 & 67 & 105\\
strength: & 440.2 & 89.5 & 14.9 & 0.5
\end{array} \]
\[\begin{array}{rcccc}
\mbox{Fig.\ref{fig:homographyFig1}b}  & red  & green & blue & cyan\\
scale: & 0.20  & 0.17 & 0.56 & 5.16\\
inliers: & 107 & 88 & 165 & 89\\
strength: & 529.6 & 517.7 & 293.4 & 17.3.
\end{array} \]

The importance of preprocessing can also be seen in \cite{serradell10},
where the geometric and appearance priors were used to increase the 
amount of consistent matches before estimation,
when the PROSAC \cite{chum05} failed in the presence of many incorrect matches.
In Fig.\ref{fig:homographyFig1}c, 
the 713 point pairs from OpenCV SIFT are used, while 
the more dense data from datasets \cite{pham14} with 1940 points
are tested in Fig.\ref{fig:homographyFig1}d for comparison.
After the estimations,
MISRE returns only two inlier structures in 
Fig.\ref{fig:homographyFig1}c,
and four inliers in Fig.\ref{fig:homographyFig1}d.
With denser inlier points, more inlier structures can be detected.

\begin{figure}[t]
\centering
\begin{tabular}{@{\hspace{-0.0cm}}c@{\hspace{0.1cm}}c}
\includegraphics[scale=0.215]{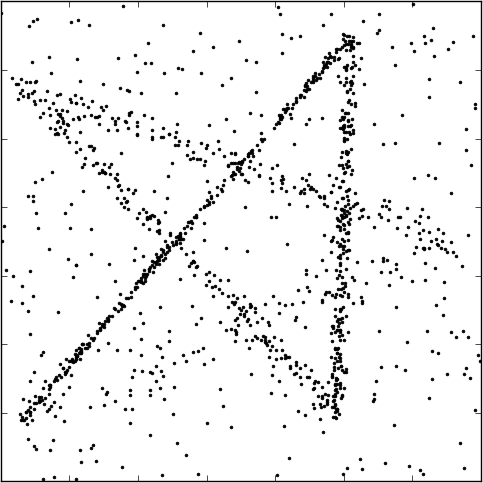}&
\includegraphics[scale=0.215]{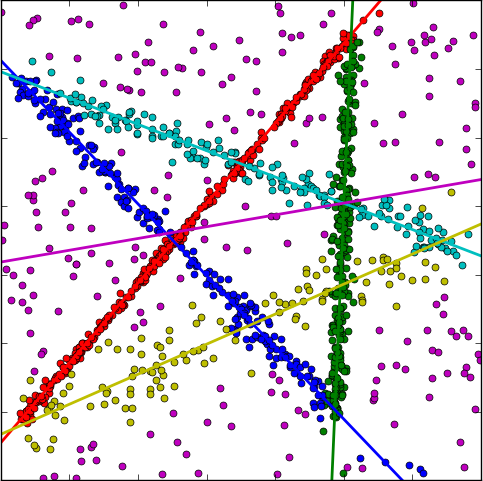} \\
(a) & (b) \\ 
\includegraphics[scale=0.335]{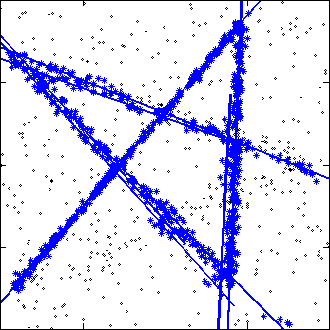}&
\includegraphics[scale=0.38]{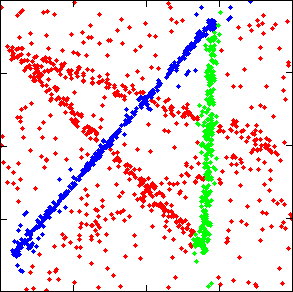} \\
(c) & (d)
\end{tabular}
\caption{Comparison of 2D lines estimations.
(a) Five synthetic lines.
(b) MISRE. (c) J-Linkage. (d) T-Linkage.
}
\label{figure_compare_line}
\vspace{-.5cm}
\end{figure}

\section{Discussion}
\label{sec:discussion}

A new robust algorithm was presented which does not require the 
inlier scales to be specified by the user prior to the estimation.
It estimates the scale for each structure adaptively from the input data,
and can handle inlier structures with different noise levels.
Using a strength based classification, 
the inlier structures with larger strengths are retained 
while a large quantity of outliers are removed. 

In Section \ref{sec:comparison} we will compare MISRE with
other robust estimators. 
In Section \ref{sec:unsolved} open problems are reviewed.

\subsection{Comparison with Other Robust Estimators}
\label{sec:comparison}
The same inlier/outlier setting in Fig.\ref{fig:2dline}a
is used in Fig.\ref{figure_compare_line}a for comparison of 2D lines
estimation. 
In Fig.\ref{figure_compare_line}b
MISRE successfully locates all the five inlier structures, 
and returns them first based on strengths.
However, as Fig.\ref{figure_compare_line}c shows,
the J-linkage fails to recover the lines without given a 
correct inlier scale for each structure, and many split lines are returned.
In Fig.\ref{figure_compare_line}d, the T-linkage cannot fully handle
the estimation with different inlier scales, and returns only two inlier structures.

\begin{figure}[t]
\centering
\begin{tabular}{@{\hspace{-0.0cm}}c@{\hspace{0.1cm}}c}
\includegraphics[scale=0.7]{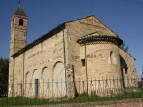}&
\includegraphics[scale=0.32]{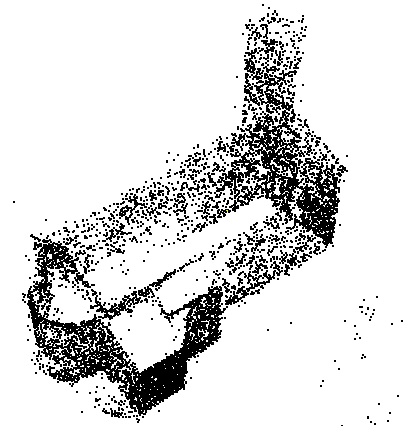} \\
(a) & (b) \\ 
\includegraphics[scale=0.3]{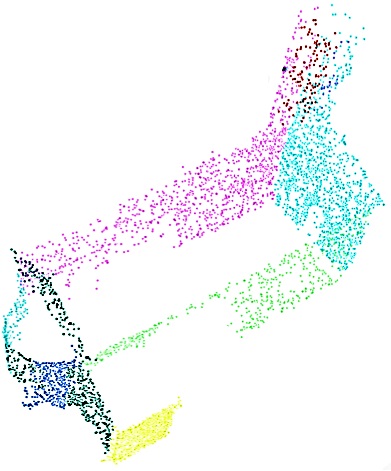}&
\includegraphics[scale=0.3]{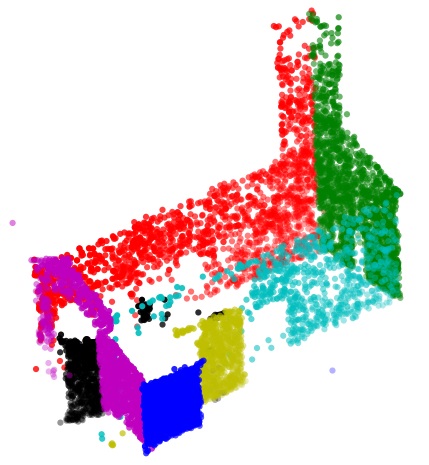} \\
(c) & (d)
\end{tabular}
\caption{Comparison of 3D planes estimations in point cloud.
(a) A sample image used in SfM algorithm.
(b) A total of 11094 points.
(c) J-Linkage.
(d) MISRE.
}
\label{figure_compare_plane}
\vspace{-.5cm}
\end{figure}

The reconstructed point cloud in \cite{Farenzena09} 
(Fig.\ref{figure_compare_plane}a) from 48 2D images
containing 11094 points in 3D were extracted (Fig.\ref{figure_compare_plane}b).
Both J-linkage and MISRE segment the point cloud into 
correct planes  in Fig.\ref{figure_compare_plane}c and
Fig.\ref{figure_compare_plane}d.
However, this estimation process takes 330 seconds in 
J-linkage's MATLAB implementation, while only 10.2 seconds in MISRE.

\begin{figure}[h]
\centering
\begin{tabular}{@{\hspace{-0.0cm}}c@{\hspace{0.1cm}}c}
\includegraphics[scale=0.235]{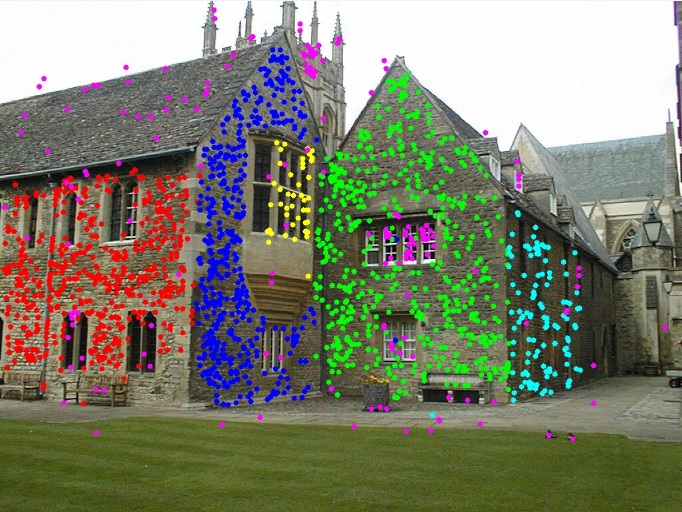}&
\includegraphics[scale=0.25]{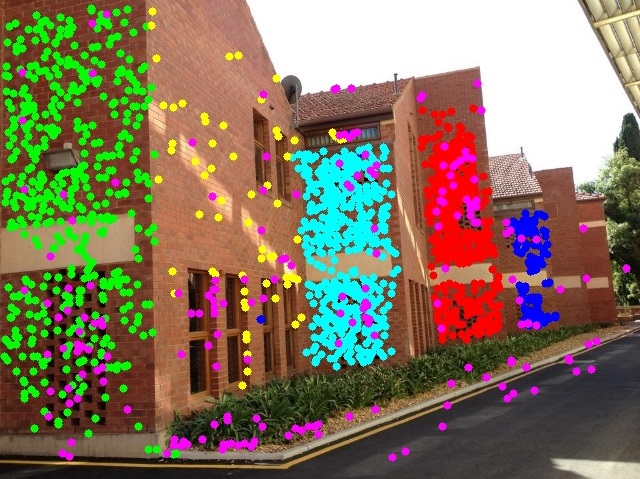}\\
(a) & (b) \\
\includegraphics[scale=0.235]{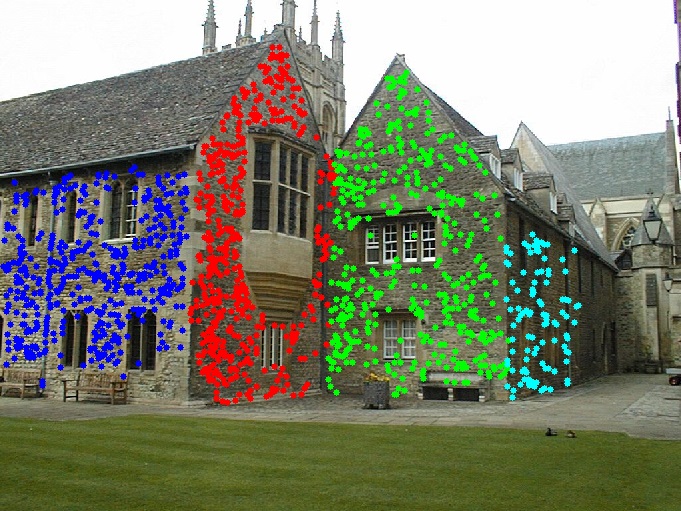}&
\includegraphics[scale=0.245]{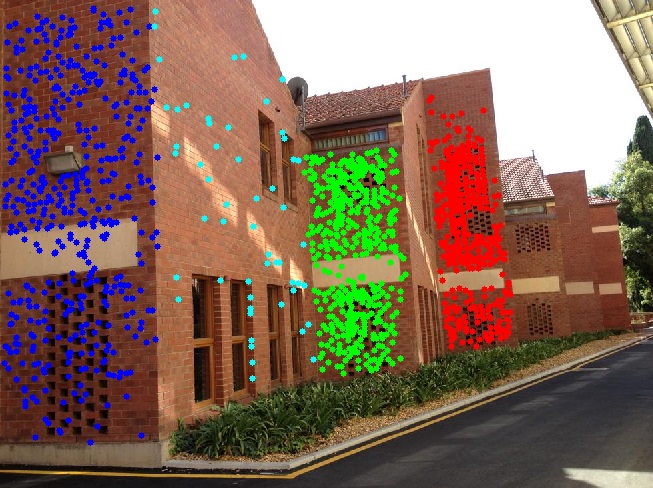}\\
(c) & (d) \\ 
\includegraphics[scale=0.25]{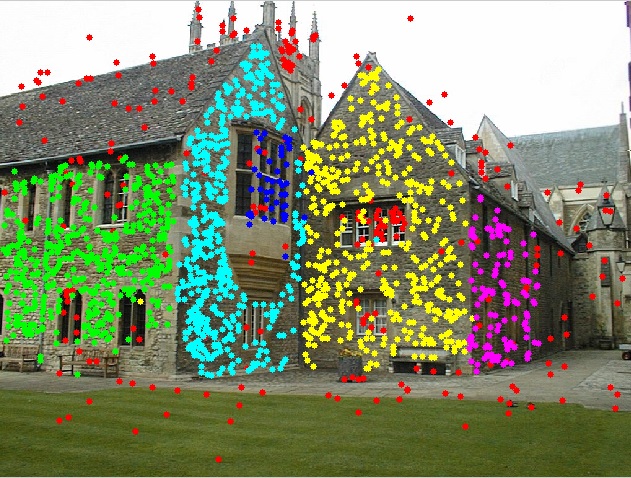}&
\includegraphics[scale=0.265]{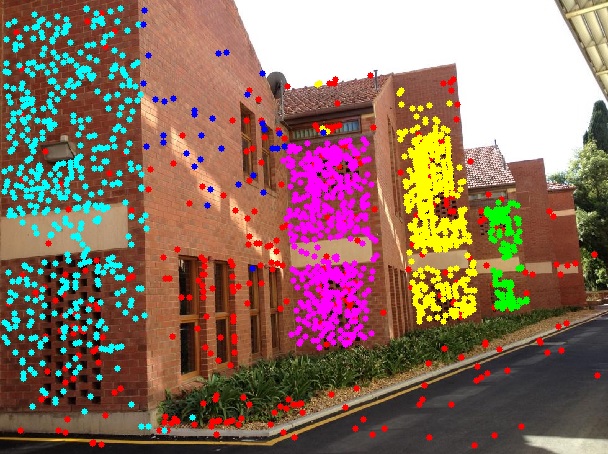}\\
(e) & (f) \\
\end{tabular}
\caption{Comparison of Homography estimations.
 Left: {\it Merton College}, 1982 poins. 
Right: {\it Unionhouse}, 2084 points.
(a)\&(b) MISRE, five inlier and one outlier structures.
(c)\&(d) gpbM, four inlier structures recovered.
(e)\&(f) RCMSA, five inlier and one outlier structures.
}
\label{fig:homographyFig2}
\vspace{-.4cm}
\end{figure}

Finally in Fig.\ref{fig:homographyFig2}, 
we show the comparison of homography estimations 
with gpbM and RCMSA.
By comparing with the ground truth, 
the correct/incorrect classification for each structure is listed
in the following table (use colors in Fig.\ref{fig:homographyFig2}a
and Fig.\ref{fig:homographyFig2}b as references):
\[\begin{array}{rccccc}
\mbox{Fig.\ref{fig:homographyFig2}a} & red  & green & blue & cyan & yellow\\
MISRE: & 491/1  & 479/9 & 499/29 & 125/0 & 53/0\\
gpbM: & 497/7 & 471/10 & 492/28 & 124/1 & N/A\\
RCMSA: & 499/1 & 485/3 & 498/8 & 133/5 & 53/9 
\end{array} \]
\[\begin{array}{rccccc}
\mbox{Fig.\ref{fig:homographyFig2}b} & red  & green & blue & cyan & yellow\\ 
MISRE: & 495/20  & 155/0 & 499/11 & 499/5 & 59/0\\
gpbM: & 495/16 & 154/1 & 498/5 & 498/1 & N/A\\
RCMSA: & 495/18 & 155/2 & 499/5 & 499/4 & 45/0 
\end{array} \]
The results obtained from these three methods are comparable.
However, since RCMSA requires tuning on the internal parameters,
the prior knowledge is always required.
MISRE significantly reduces the processing time (in seconds)
\[\begin{array}{rcc}
\mbox{Proccessing Time} & \mbox{Merton College} & \mbox{Unionhouse}\\
MISRE: & 3.07  & 3.78\\
gpbM: & 477 & 495\\
RCMSA: & 24.58 & 25.40 
\end{array} \]
Both gpbM and RCMSA take much longer processing time in their
respective estimation process, which again verifies
the efficiency of the new estimator MISRE.

\vspace{-.2cm}
\subsection{Open Problems}
\label{sec:unsolved}

We will list several open problems, where further research and experiments are still needed.

Assume that the measurements of the input points
have different covariances which are not specified.
In many computer vision problems this situation 
is neglected but can still exist. 
The homoscedastic inlier covariances have the form
$[ \sigma_1^2 \ldots \sigma_{\scriptsize{\by}}^2] \bI_{\scriptsize{\by}}$.
with $\sigma_j$-s unknown. 
The computation for the covariance of the carrier
(\ref{eqn:newcovariance}) places the $\sigma_j$-s
into the product of two Jacobian matrices.
A possible solution is to start with a uniform $\sigma$, 
that is, $\sigma^2\bI_{\by}$. 
Each inlier structure may not attract the 
quasi-correct amount of points. 
Then for each inlier structure separately,
consider a local region where the inliers are located.
Apply the entire estimation again only on the data inside this region,
in this way a more accurate $\hat{\sigma}_j$ could be found and 
to update the final estimate.

In face image classification or projective motion factorization,
the objective functions have only one carrier vector $\zeta = 1$, but the
estimate is an $m\times k$ matrix $\bTheta$ and 
a $k$-dimensional vector $\balpha$. 
Since $\zeta$ is one dimensional, here we
use $\bx$ instead of $\tilde\bx$.
The covariance of $\bz_i$ is
$\sigma^2\bH_i = \sigma^2\bTheta^\top \bC_i \bTheta$, 
with $\sigma^2$ unknown.
This gives a $k\times k$ symmetric Mahalanobis distance matrix 
for $i=1,\ldots,n$
\begin{equation}
\label{eqn:matrix}
\bD_i = \sqrt{ \left( \bx_i^\top \bTheta - 
\balpha \right)^\top \bH_i^{-1} 
\left( \bx_i^\top \bTheta - \balpha \right)}
\end{equation}
which could be expressed as the union of $k$ vectors 
$\bD_i = [\bd_{i:1}~\ldots~ \bd_{i:k}]$ .
A possible solution is to 
order the Mahalanobis distances $\bd_{[i:*]}$
for each column separately,
and collect the inputs corresponding to the minimum sum of 
distances for $\epsilon\%$ of the data.
The $k\times k$ matrices are reduced to $k$ initial sets, 
one for each dimension. 

Apply independently $k$ times the
expansion process described in Section \ref{sec:newsigma} and define the 
$k\times k$ diagonal scale matrix with $\hat{\sigma}_j, ~~ j=1,\ldots,k$.
The $k\times k$ covariance matrix is computed as
$\bB_i = \bS_k^\top \, \bTheta^\top \bC_i \bTheta \, \bS_k .$
The second step in the algorithm, the mean shift, is now multidimensional and further experiments will be needed to verify 
the feasibility of this solution.

If an image contains, say, both planes and spheres, 
there is no clear way to estimate both of them properly.
See for example, Fig.\ref{figure_7}.
Supposing we start with the planes, 
then some points from the spheres 
may be misclassified as planes. 
These points should be 
put back into the input data for another sphere estimation,
otherwise some spheres may not be detected. 
The same problem arises when we estimate the spheres first.
The correct separation of multiple types of structures
could require supplemental processing.

The Python/C++ program for the robust estimation of 
multiple inlier structures is posted on our website at \\
{\tt \centerline{rci.rutgers.edu/riul/research/code/}
\centerline{MULINL/index.html.}}

\bibliographystyle{alphaieeetr}
\bibliography{ourreference}

\end{document}